\documentclass[letterpaper,twocolumn,10pt]{article}
\usepackage{usenix}
\usepackage{my_package}

\usepackage{tikz}
\usepackage{amsmath}
\usepackage{comment}

\begin{document}

\date{}

\title{\Large \bf Beyond Heavy Log Curation: Perplexity-Based APT Detection \\via Unsupervised, Context-Augmented Language Models}

\author{
  \textrm{Shoya Otsu}\textsuperscript{1},
  \textrm{Kei Suzuki}\textsuperscript{2},
  \textrm{Toshiaki Koike-Akino}\textsuperscript{2},
  \textrm{Jing Liu}\textsuperscript{2},
  \textrm{Ye Wang}\textsuperscript{2},
}

\maketitle

\footnotetext[1]{
    Mitsubishi Electric Corporation, Japan.\\
    \makebox[2.5em][r]Email: \texttt{Otsu.Shoya@eb.MitsubishiElectric.co.jp}}
\footnotetext[2]{Mitsubishi Electric Research Laboratories, Cambridge, MA, USA.}

\begin{abstract}
Advanced Persistent Threats (APTs) remain difficult to detect because only a small fraction of events in large-scale logs are attack-related, and investigation is expensive and hard to scale. Prior machine-learning approaches can reduce analyst workload, but they often rely on heavily curated training data and sophisticated preprocessing pipelines. Building and maintaining such pipelines require substantial domain expertise and engineering cost. Motivated by insights from a study of a strong APT detection baseline, we propose CAPTAIN (Context-Augmented Perplexity-based Threat Activity log detectIoN), a perplexity-based detector that leverages general, pre-trained language models with minimal, domain-agnostic preprocessing, enabling robust scoring of long, minimally processed log entries. CAPTAIN encodes recent history with an encoder model and a Q-Former-style bridge, then injects the compact context tokens into the decoder input so that perplexity reflects temporal context. To improve stability, CAPTAIN additionally applies smoothing filters to the perplexity time series. Across APT-oriented benchmarks, CAPTAIN competes with strong existing baselines and remains robust under substantially less curated inputs, that reduces the development and operational cost of advanced log preprocessing.
\end{abstract}

\section{Introduction}

\subsection{Background}
\begin{figure*}[t]
\centering
\includegraphics[width=\textwidth]{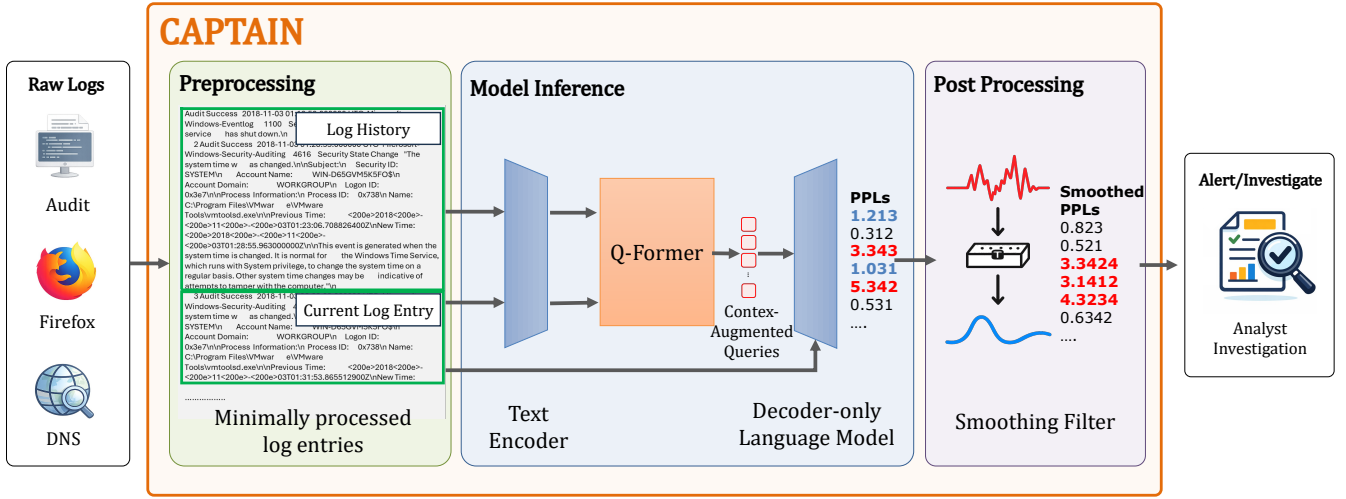}
\caption{\label{fig:comp_pipe} \textbf{CAPTAIN overview.} CAPTAIN ingests minimally processed, time-ordered logs (the current entry and its recent history). A text encoder summarizes the history, and a Q-Former bridges it into a decoder-only LM as context tokens. The LM assigns a perplexity (PPL) score to each entry. Then, the PPL time series is smoothed in post-processing, and entries with high final scores are flagged as suspicious for analyst investigation.}
\end{figure*}

Advanced Persistent Threats (APTs) remain a major operational challenge for defenders because they unfold over long time horizons and blend into benign activity.
In practice, defenders must examine massive volumes of heterogeneous logs, and only a small fraction of events are truly attack-related.
Manual processes are costly and difficult to scale. Therefore, machine learning applications for attack detection are being actively investigated as a way to automate these processes.

Recent advances in machine learning and natural language processing, especially the emergence of Transformer~\cite{vaswani2017attention}, have greatly improved text understanding, summarization, generation, and reasoning~\cite{ lewis2020bart, wei2022cot, raffel2020exploring}.
Encoder-style transformers such as BERT~\cite{devlin2019bert} use bidirectional attention to compress token-level information into rich representations.
They achieve strong performance on many language understanding tasks, including tasks that require document-level representations.
In contrast, decoder-only transformers such as the GPT family~\cite{brown2020language,ouyang2022training} and Qwen~\cite{yang2025qwen3} compute next-token probabilities in an autoregressive manner.
This design enables high-quality text generation and general-purpose reasoning .
Transformer-based models have also been widely adopted for log analysis and anomaly detection \cite{guo2021logbert, han2023loggpt, lee2023lanobert,lin2016logclustering, du2017deeplog}.

A common approach in recent attack investigation starts from audit logs and constructs a provenance or causal graph.
The graph connects system entities and events through dependency edges, and it helps analysts trace multi-step attack paths.
Many systems then apply machine learning on top of the graph to identify suspicious nodes or subgraphs and to reconstruct the attack story \cite{jia2024magic,cheng2023kairos,lv2024trec}.
ATLAS \cite{alsaheel2021atlas} is a representative benchmark and method in this line of work.
It builds a causal graph from audit logs and learns patterns of attack and non-attack behaviors for investigation.
The shared evaluation setting of ATLAS has enabled rapid progress and comparison across methods in the community.

However, this graph-based workflow raises practical challenges.
First, graph construction requires heavy preprocessing, which adds latency and engineering overhead.
Second, many learning-based methods benefit from labels on graph nodes or edges, but reliable labeling is costly and error-prone.
Third, graph structures and schemas often depend on the dataset and the collection environment, which limits portability across deployments.
Finally, investigation pipelines often focus on post-hoc analysis.
In operations, defenders also want fast detection on each host, not only centralized aggregation and long offline analysis.

To reduce the burden on analysts, recent research has explored machine-learning-based approaches that automatically surface suspicious events and prioritize investigations.
AIRTAG \cite{ding2023airtag}, for example, reports strong detection capability on the ATLAS dataset and avoids manual graph labeling by learning directly from log text.
Despite this progress, existing approaches still depend heavily on dataset-specific preprocessing and curation.
Such dependence increases development cost and can limit robustness when logs are long and noisy.

In this work, we revisit AIRTAG as a strong baseline and conduct a replication study to better understand the factors driving its performance.
We then propose a new approach that better matches practical logging conditions and addresses the limitations identified in our replication study.

\subsection{Revisiting AIRTAG: Reproduction Findings and Practical Limitations}
Through our replication study, we identified two factors that substantially affect measured performance, as well as two practical limitations that become more prominent in realistic logging settings:

\begin{itemize}
    \item \textbf{Observation 1: Cross-split correlation of malicious log entries can introduce evaluation bias.}
    Malicious log entries in ATLAS can be highly correlated across splits, sometimes appearing in near-identical form in both training and test sets. 
    This overlap can lead to optimistic estimates by favoring memorization over generalization, so we re-evaluate under a setting that reduces such overlap.

    \item \textbf{Observation 2: Post-processing can leverage malicious log entry labels.}
    AIRTAG includes a post-processing filter that can utilize labels that indicate malicious log entries.
    We find that removing these labels changes the behavior of post-filtering and leads to degraded performance.

    \item \textbf{Limitation 1: Reduced robustness to long log entries and reliance on manual feature extraction.}
    AIRTAG’s log preprocessing pipeline involves carefully designed manual feature extraction that may be key to enable attack detection
    from short token sequences.
    In our experiments, however, we found that AIRTAG’s performance degrades once the token budget is increased beyond a certain point.
    We also observed performance drops when using a minimally processed dataset that does not rely on careful application of domain knowledge for manual feature extraction.
    These results suggest that AIRTAG is more sensitive to token content and
    may require high-quality, domain-specific preprocessing to achieve strong performance in operational settings.

    \item \textbf{Limitation 2: Lack of context modeling across log events.}
    AIRTAG evaluates each log entry independently as benign or malicious, and therefore cannot naturally incorporate temporal context or dependencies across events. This is particularly important for detecting APTs (advanced persistent threats): because APT campaigns unfold over long time horizons, attacker activity may be temporally sparse or occur in bursts, and the sequence and timing of events can provide crucial evidence that is not captured by individual log lines alone.
\end{itemize}

\subsection{Our Approach}
In this paper, we propose CAPTAIN (Context-Augmented Perplexity-based Threat Activity log detectIoN), a perplexity-based detector that utilizes general, pre-trained language models with minimal, domain-agnostic preprocessing (Figure~\ref{fig:comp_pipe}).
By leveraging LMs, the method can judge whether a token sequence is plausible under normal behavior for both short and long log entries. This allows us to ingest log entries with minimal, domain-agnostic preprocessing while preserving most of the information in raw logs, which is important because attack evidence may appear anywhere in a long message.
To incorporate past log context efficiently, we create a small number of special context tokens using an encoder-style Transformer and a Q-Former module \cite{li2023blip2} that bridges the representation spaces of the encoder and the decoder LM. We insert these tokens into the LM input so that the LM can consider historical activity when scoring the current log entry.
Finally, we treat the sequence of perplexity scores as a time series and apply a classical post-filtering step (e.g., smoothing), which improves stability and makes sustained abnormal behavior easier to detect. We train the model with an autoencoder-style unsupervised learning scheme so that it can learn normal patterns without attack labels.

This paper makes the following contributions:
\begin{itemize}
    \item \textbf{Re-evaluation of AIRTAG.}
    We reproduce AIRTAG and identify artifacts that can unintentionally introduce label-related information into the learning/evaluation pipeline.
    By correcting these issues and re-evaluating, we clarify AIRTAG's performance under conditions that better reflect deployment settings.

    \item \textbf{Context-aware detection model with only minimal domain-specific preprocessing.}
    We propose CAPTAIN, a new approach that incorporates historical log context while operating on logs produced by simple, domain-agnostic preprocessing rather than carefully curated feature extraction.
    Concretely, we show that soft prompting of contextual log information via a Q-Former bridge is effective for attack detection, and ablation results confirm that the gain cannot be explained by domain adaptation alone.
    
    \item \textbf{Score smoothing improves performance.}
    We find that treating per-entry perplexity scores as a time series and applying a smoothing filter consistently improves detection performance and stabilizes the detector.
    
    \item \textbf{Unsupervised learning formulation for our proposed model.}
    We develop an unsupervised training method that learns the proposed context-conditioning mechanism using only benign logs, enabling CAPTAIN to be trained without relying on labeled malicious examples.

\end{itemize}

\section{Establishing a Baseline: AIRTAG}
We conducted a replication study of AIRTAG to establish a fair and well-controlled baseline for CAPTAIN.
In the course of this effort, we identified several evaluation-related issues that we believe are important to document and share, and we summarize them here.

\subsection{Preprocessing}
\paragraph{Dataset.}
AIRTAG evaluates its method on several datasets: the public ATLAS~\cite{alsaheel2021atlas} (M1–M6 and S1–S4), DEPIMPACT~\cite{pengcheng2022backpropagation} (dataleak, vpnfilter, and shellshock), and an additional dataset collected by the authors. 
However, the scripts and detailed results needed to reproduce AIRTAG's evaluation on DEPIMPACT and the privately collected dataset are not available~\cite{abrar2025onthereproducibility}, making an exact reproduction difficult.
Therefore, in this work, we focus on the reproducible, publicly available ATLAS dataset.

\paragraph{Preprocessing.}
In principle, AIRTAG can preprocess logs in various formats. In the implementation released on GitHub, however, the evaluation on ATLAS relies on data that has already been preprocessed using the ATLAS preprocessing procedure. This procedure aligns logs from multiple sources—such as Firefox, DNS, and security/audit events—by timestamp and merges them into a single stream.
Concretely, ATLAS parses three log types (Firefox, DNS, and Audit logs) and converts each log entry into a CSV row with 19 fields. The last field contains both (i) a source identifier (e.g., ``LA'', ``LB'', or ``LD'' to respectively distinguish Firefox, DNS, or Windows Audit logs), and (ii) the ground-truth label: benign indicated as ``-'' or malicious as ``+''.

Based on the ATLAS-preprocessed data, AIRTAG performs additional preprocessing to generate the training split, the test split, and a vocabulary file for its custom tokenizer.
During our replication, we confirmed that the BERT training and test files provided by the AIRTAG authors still contained the malicious markers ``+'' at the end of each line. Since this marker correspond to the ground-truth label, we remove it from both the training and test data in our experiments to avoid unintentionally exposing label information to the model.

\subsection{Training}
In the training stage, first, AIRTAG tokenizes log entries using a domain-specific tokenizer and pretrains BERT (uncased\_L-6\_H-128\_A-2) on a mixture of benign and malicious log entries.
After pretraining, it encodes the benign-only training set into embedding vectors and trains a downstream One-Class Support Vector Machine (OC-SVM) using these benign embeddings.
A key property of ATLAS is that malicious events were synthetically created in a manner that resulted in malicious log entries that share distinguishing features across all dataset splits.
Specifically, malicious log entries are uniquely identified by containing keywords \texttt{0xalsaheel.com}, \texttt{192.168.223.3} or \texttt{payload.exe}, which,
as shown in Table~\ref{tab:malicious-keywords}, are shared across all data splits. 

This raises the risk that pretraining may expose the model to test-set patterns, which can inflate evaluation results. To reduce this risk in our replication, we modify the training data so that malicious keywords that appear in the training split do not exactly match those in the test split. Concretely, we replace malicious keywords in the training split with alternative ones, as listed in Table~\ref{tab:keyword_mapping}, ensuring that only the test split contains original malicious keywords.
\begin{table}[t]
  \centering
  \caption{The number of log entries containing common malicious keywords, across all ATLAS dataset splits.}
  \label{tab:malicious-keywords}

  \resizebox{\linewidth}{!}{%
  \begin{tabular}{lrrrrrrrrrr}
    \toprule
    Keyword & M1 & M2 & M3 & M4 & M5 & M6 & S1 & S2 & S3 & S4 \\
    \midrule
    0xalsaheel.com   & 51 & 89 & 113 & 13 & 62 & 16 & 71 & 41 & 12 & 19 \\
    192.168.223.3    & 342 & 780 & 623 & 355 & 553 & 2506 & 301 & 247 & 1363 & 413 \\
    payload.exe      & 8940 & 34956 & 34963 & 9057 & 34037 & 9071 & 5162 & 17855 & 4422 & 17862 \\
    \bottomrule
  \end{tabular}%
  }
\end{table}

\begin{table}[t]
  \centering
  \caption{Malicious Keyword Mapping (left $\rightarrow$ right)}
  \label{tab:keyword_mapping}
  \begin{tabular}{ll}
    \hline
    Before (Original) & After \\
    \hline
    0xalsaheel.com & 0xygen.org \\
    192.168.223.3 & 192.168.7.58 \\
    payload.exe & shipment.exe \\
    pypayload.exe & pyshipment.exe \\
    msf.rtf & hghg.rft \\
    msf\_2018\_8174.rtf & fgfg\_1990\_0605.rft \\
    msf.exe & fghg.exe \\
    msf.doc & hgfg.doc \\
    aalsahee/index.html & 00xygen/start.html \\
    \hline
  \end{tabular}
\end{table}

\subsection{Attack Detection}

During inference, AIRTAG applies the pretrained BERT encoder and the OC-SVM to determine whether each log entry is likely related to an attack.
Log entries flagged as suspicious are then passed through a frequent-term-based post-filter (referred to as field frequency filters with tolerance bounds) to reduce false positives rate (FPR). 
While inspecting the implementation of this post-filter, we found that the file used for word counting still contains the ground-truth malicious/benign labels ``+/-''.
In our analysis, we discovered that the presence of these labels play a crucial factor in the effectiveness of this filter, by observing that removing these ``+/-" labels significantly degraded performance.
demonstrating that it no longer performed its intended function.
Therefore, we do not use this post-filter in the remainder of this paper.
Details of this analysis is described in the Appendix~\ref{apx:post_filter}.

\section{Proposed Method}

\begin{figure*}[t]
\centering
\includegraphics[width=\textwidth]{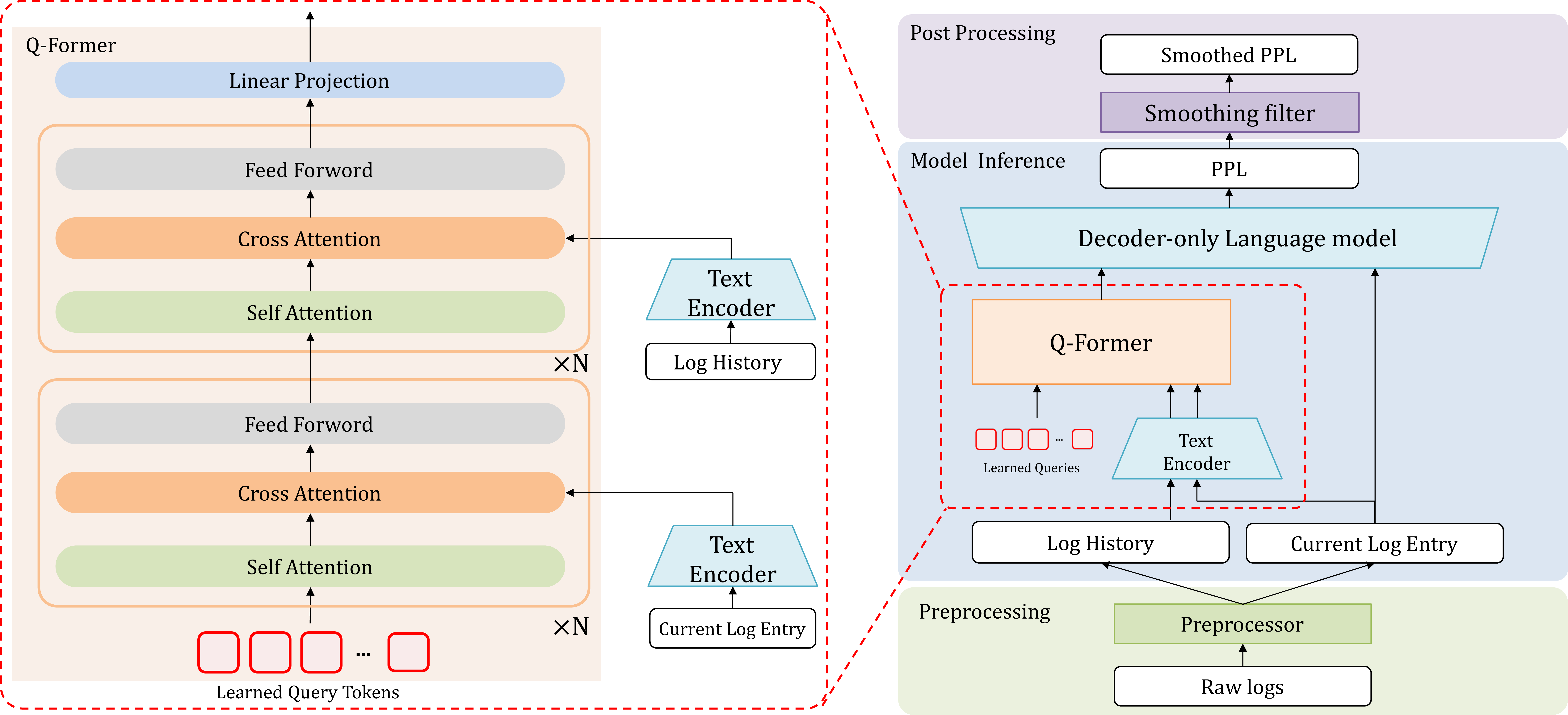}
\caption{\label{fig:model_arch} \textbf{CAPTAIN architecture with Q-Former details.} The text encoder embeds the current log entry and log history. Q-Former uses a fixed set of learned query tokens and updates them with Transformer blocks (self-attention, cross-attention to encoder outputs, and feed-forward layers). In our two-stage design, queries are first conditioned on the current entry and then refined using the log history, followed by a linear projection into the decoder LM space. The resulting context tokens are fed to a decoder-only LM to compute per-entry perplexity (PPL), and the PPL time series is smoothed in post-processing to produce stable anomaly scores.}
\end{figure*}

CAPTAIN is a perplexity-based attack detection method that incorporates historical log context with minimal, domain-agnostic preprocessing.
The pipeline has three stages: preprocessing, model inference, and post-processing (Figure \ref{fig:model_arch}).

\subsection{Perplexity-based Detection of Attack-related Log Entries}

In this study, we use perplexity as a measure of how well an LM can explain each log entry as a ``normal'' log, while explicitly conditioning on historical log context.
Let a current log entry be represented as a token sequence $\mathbf{x}=(x_1,\ldots,x_T)$, and let $\mathbf{c}$ denote the context provided to the LM (e.g., context tokens derived from past logs and inserted into the LM input).
An autoregressive LM defines the conditional likelihood of $\mathbf{x}$ given $\mathbf{c}$, as a product of next-token conditionals:
\begin{equation}
q(\mathbf{x}\mid \mathbf{c}) = \prod_{t=1}^{T} q(x_t \mid x_{<t}, \mathbf{c}),
\end{equation}
where $x_{<t} \triangleq (x_1,\ldots,x_{t-1})$. We then define the average cross entropy of the sequence conditioned on $\mathbf{c}$ as
\begin{equation}
H(\mathbf{x}\mid \mathbf{c}) \triangleq -\frac{1}{T}\sum_{t=1}^{T}\log q(x_t \mid x_{<t}, \mathbf{c}),
\end{equation}
and the conditional perplexity as the exponential of this cross entropy:
\begin{equation}
\mathrm{PPL}(\mathbf{x}\mid \mathbf{c}) \triangleq \exp\!\bigl(H(\mathbf{x}\mid \mathbf{c})\bigr).
\end{equation}
A larger $\mathrm{PPL}(\mathbf{x}\mid \mathbf{c})$ indicates that the model finds the sequence harder to predict (i.e., it has lower likelihood).

Our key assumption is that attack-induced logs differ from normal logs in terms of vocabulary, syntax, and transitions of log context; therefore, an LM trained on normal logs is likely to assign lower probabilities to malicious logs, resulting in higher perplexity.
Accordingly, we compute perplexity of each log entry and decide it is an attack, if it exceeds a threshold $\tau$:
\begin{equation}
\mathbf{x}\ \text{is suspicious} \iff \mathrm{PPL}(\mathbf{x}\mid \mathbf{c}) > \tau.
\end{equation}

\subsection{Preprocessing}

CAPTAIN is designed to minimize the need for domain-specific preprocessing.
Rather than relying on manual feature extraction (e.g., regex-based parsing), we apply only lightweight, general transformations that keep the log content largely intact while making it suitable as text input.
Each event is associated with a timestamp, and when a single event spans multiple lines, we merge the lines into a single textual record.
Specifically, we normalize raw audit log entries by converting timestamps to UTC and replacing line breaks with the string ``\textbackslash n'', so that a single multi-line Audit event is represented as a single-line textual record.

We also follow the dataset conventions of ATLAS for DNS and Firefox log types: for DNS logs, we retain only responses because the corresponding request information is already reflected in the response; for Firefox logs, we focus on events related to HTTP requests. Importantly, these choices do not require designing log-specific feature templates, but instead avoid heavy curation while preserving the original semantics of the logs.

CAPTAIN does not assume a specific internal schema of the log entry. As long as the input can be represented as text such that the model can access both (i) the current log entry to be scored and (ii) the historical log context preceding it, CAPTAIN can be applied without requiring a fixed format.

\subsection{Model}

As shown in Figure~\ref{fig:model_arch}, CAPTAIN has three model components: (i) a text encoder, (ii) a decoder-only transformer LM, and (iii) a Q-Former that bridges the representation spaces of the encoder and the LM.

\textbf{Text encoder.}
The text encoder converts both the current log entry (the entry we want to score) and the past log entries (context) into continuous embeddings.

\textbf{Q-Former bridge.}
Q-Former was originally introduced in BLIP-2~\cite{li2023blip2} as a lightweight bridge module that projects encoder-side embeddings into a form that a downstream LM can effectively use. It maintains a fixed number of learned query vectors and extracts information from external encoder embeddings through cross-attention. Through training, Q-Former learns (1) how to select relevant information from the encoder outputs and (2) how to map that information into an embedding space that the LM can interpret.

In CAPTAIN, we designed a two-stage Q-Former:
\begin{itemize}
    \item \textbf{Stage 1:} The Q-Former encodes the current log entry into the learned queries, producing a compact representation that captures what the model should focus on for this specific entry.
    \item \textbf{Stage 2:} Conditioned on the Stage-1 queries, the Q-Former then encodes the previous $C$ log entries into updated queries. This design encourages the model to extract context from past logs that is strongly related to the current entry, rather than producing a generic summary of history.
\end{itemize}

The final learned queries are inserted into the decoder LM as soft context tokens, enabling the LM to compute perplexity for the current entry while taking past context into account.

\textbf{Model choices in this paper.}
CAPTAIN does not require specific backbone models for the text encoder or the decoder LM. In our experiments, we use DistilBERT~\cite{sanh2019distilbert} as the text encoder and Qwen3-0.6B~\cite{yang2025qwen3} as the decoder LM. We intentionally choose relatively lightweight models because we target a practical use case where detection is performed quickly on each host with limited resources, rather than aggregating logs to a central server and relying on large-scale compute.

\subsection{Post Processing}

In the post-processing stage, we treat the PPL values produced by the LM as a time-series random variable and apply smoothing to clean out fluctuations.
APT activity is often temporally uneven; consequently, attack traces in logs can exhibit temporal concentration. 
Our analysis suggests that ATLAS exhibits a similar tendency; we provide a detailed characterization in the Appendix~\ref{apx:temporal_bias}.
Based on this observation, we consider the sequence of perplexity scores as a time series following the hidden Markov model. 
For such signals, smoothing with a classical filter (e.g., a moving average) can significantly reduce local noise, improve the stability of the detector, and make sustained abnormal behavior easier to identify.

\textbf{Wiener filter for smoothing.} 
In this paper, we apply a finite-impulse-response (FIR)-based \emph{smoothing filter} to the raw PPL score sequence in order to suppress short-term fluctuations and emphasize the underlying trend. 
Let $\{x_t\}_{t=0}^{T-1}$ denote the (optionally rescaled) scalar PPL score at time index $t$. 
We produce a smoothed sequence $\{\tilde{x}_t\}$ by linear filtering.

For a window length $L$ (odd), the smoothed score is computed by FIR convolution:
\begin{equation}
\tilde{x}_t = \sum_{i=0}^{L-1} w_i \, x_{t+i-c},
\end{equation}
where $\mathbf{w} = [w_0,\ldots,w_{L-1}]^\top$ is the Wiener filter coefficient vector, and 
$c = \frac{L-1}{2}$
is the center index.
The filter coefficients are obtained by solving a Wiener--Hopf system:
\begin{equation}
\mathbf{R}\mathbf{w} = \mathbf{r},
\end{equation}
where $\mathbf{R}\in\mathbb{R}^{L\times L}$ is a symmetric Toeplitz auto-correlation matrix and $\mathbf{r}\in\mathbb{R}^{L}$ is a cross-correlation vector with respect to the window center. 
Assuming a simplified Markov chain, we parameterize the correlation structure using a \emph{crossover probability} $p\in(0,1)$ as follows:
\begin{align}
R_{ij} &= (1-p)^{|i-j|} + \sigma^2 \delta_{ij}, \qquad i,j=0,\ldots,L-1, \label{eq:wiener_R}\\
r_i &= (1-p)^{|i-c|}, \qquad i=0,\ldots,L-1, \label{eq:wiener_r}
\end{align}
where $\delta_{ij}$ is the Kronecker delta and $\sigma^2 \ge 0$ is a \emph{noise variance} (diagonal loading) that regularizes the inversion of $\mathbf{R}$ and controls smoothing strength.
The FIR filter could be normalized as $\mathbf{w} / \|\mathbf{w}\|$ to preserve the signal energy, however it has no impact on final detection performance with a correspondingly scaled threshold.

\textbf{Parameter roles.}
The window size $L$ determines the temporal span of smoothing: larger $L$ yields stronger averaging but may blur rapid transitions. The crossover probability $p$ controls the correlation decay $(1-p)^{|k|}$: smaller $p$ implies longer-range correlation and typically produces a broader, smoother kernel. 
The variance $\sigma^2$ acts as diagonal regularization; increasing $\sigma^2$ reduces sensitivity to the assumed correlation model and tends to produce more conservative (less peaky) filter weights.
Note that the Wiener filter reduces to the moving average filter, $\mathbf{w}=\tfrac{1}{1+L/\sigma^2} \mathbf{1}$, when zero crossover probability is assumed (i.e., $p=0$).

\subsection{Model Training}

\begin{figure}[t]
\centering
\includegraphics[width=\columnwidth]{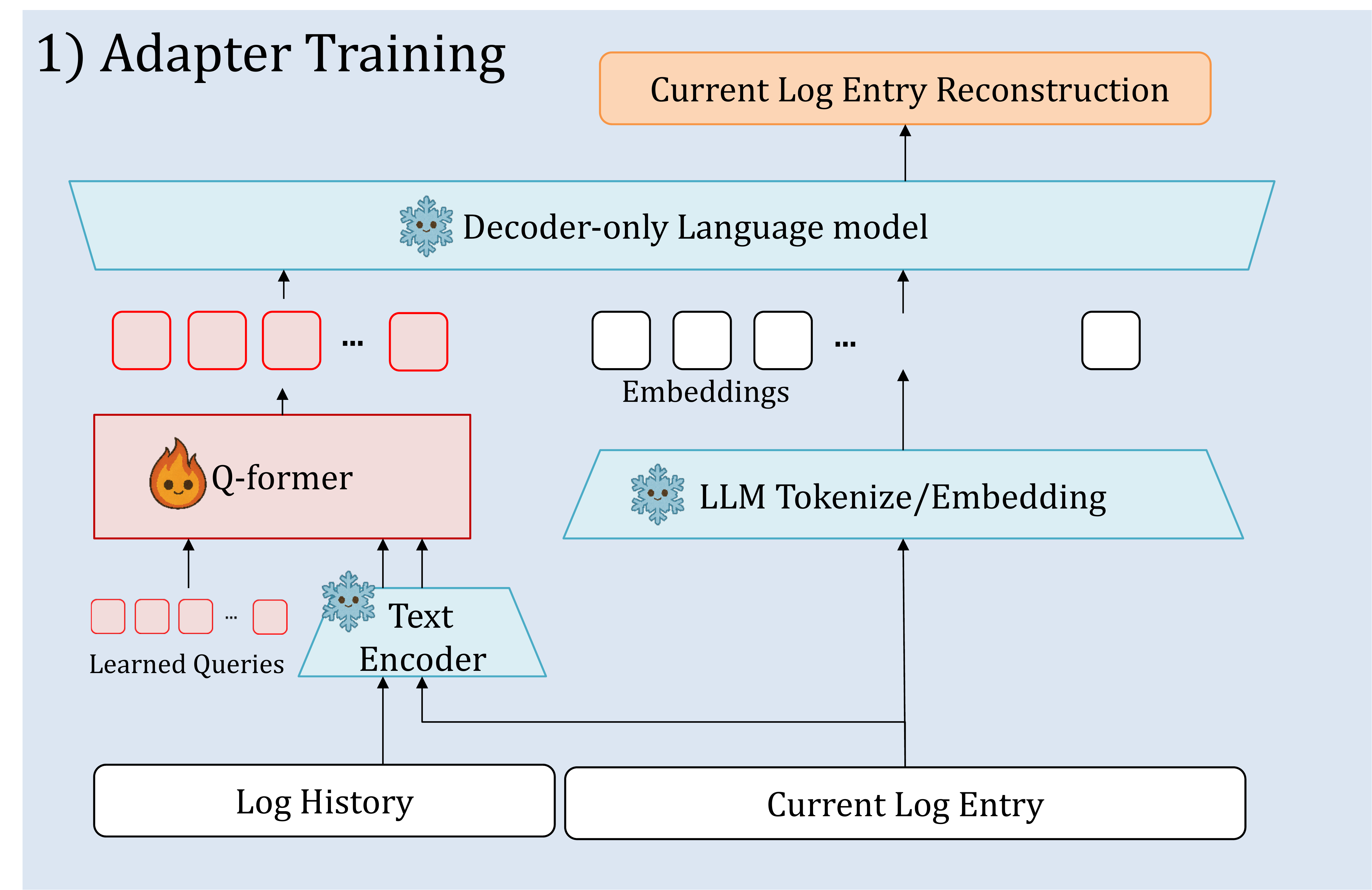}
\caption{\label{fig:stage1-training} \textbf{Stage 1:}
Training only Q-Former adapter module.
}
\end{figure}

\begin{figure}[t]
\centering
\includegraphics[width=\columnwidth]{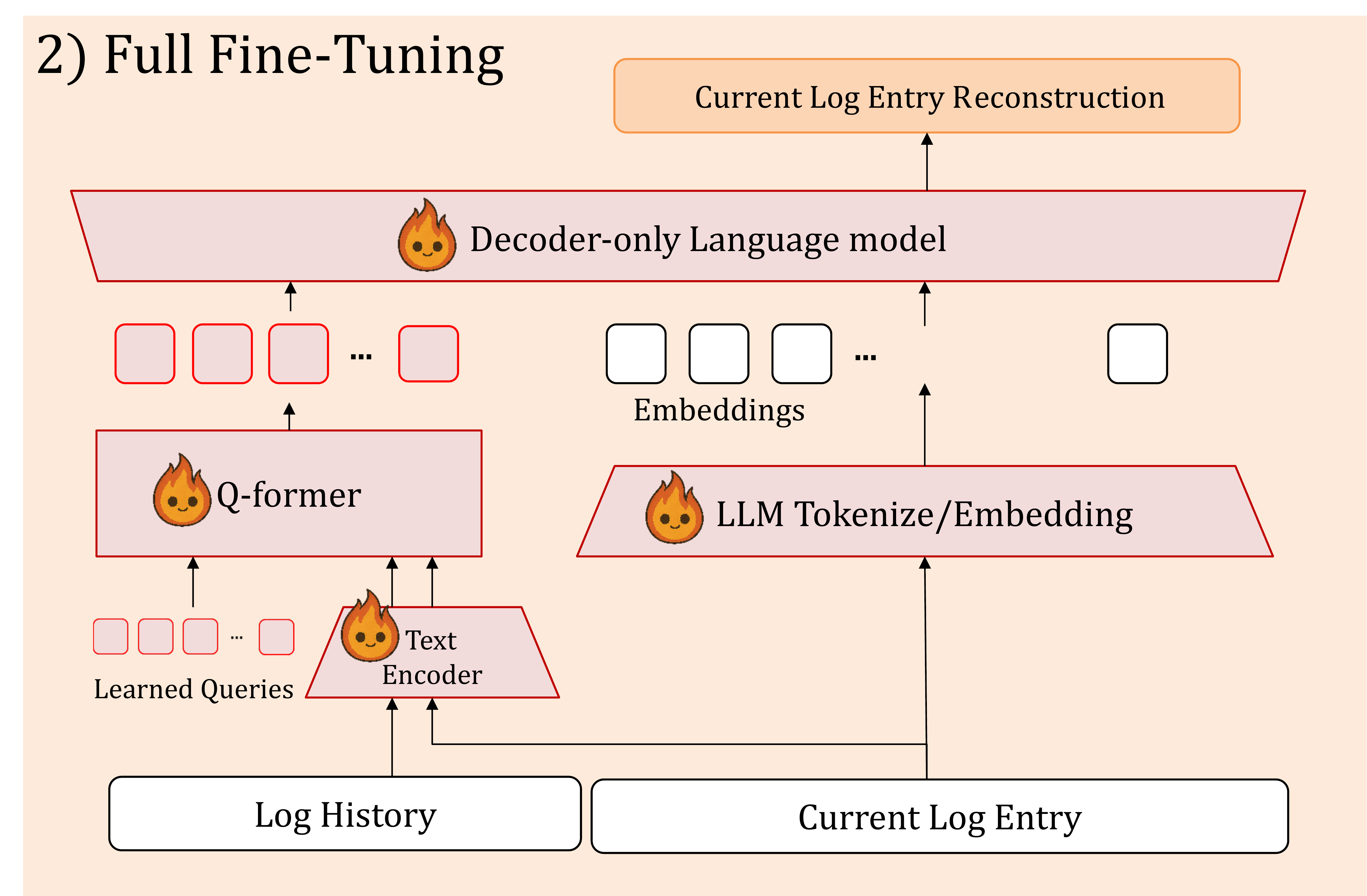}
\caption{\label{fig:stage2-training} \textbf{Stage 2:}
Full fine-tuning of all modules.
}
\end{figure}

Training in CAPTAIN consists of two stages: \emph{adapter training} (Figure~\ref{fig:stage1-training}) and \emph{full fine-tuning} (Figure~\ref{fig:stage2-training}).

\paragraph{Adapter training.}
The first stage is to learn the Q-Former’s ability to project information from the encoder space into a representation that is usable by the decoder LM.
We adopt an autoencoder-style unsupervised objective, where the model is trained to reconstruct the LM input from its own output.
During this stage, we freeze the parameters of the text encoder and the LM, and update only the parameters of the Q-Former.
The Q-Former is randomly initialized.

We use the following training configuration: 1 epoch, 100K steps, and batch size of 2, resulting in a total of 200K benign log entries used for training.
We use AdamW with a learning rate of $1\times 10^{-4}$, a weight decay of $0.01$, $\beta_1=0.9$, $\beta_2=0.98$, and 6{,}000 warm-up steps.
For reference, adapter training takes approximately 9 hours on a single NVIDIA A40 GPU (48\,GB).

\paragraph{Full fine-tuning.}
In the second stage, we fine-tune \emph{all} model components, including the text encoder, the pretrained Q-Former, and the decoder LM.
We use the same autoencoder-style unsupervised objective as in adapter training (i.e., reconstructing the LM input).
We train for 1 epoch, 30K steps, and batch size of 2, corresponding to a total of 60K benign log entries.

We use AdamW with a weight decay of $0.05$, $\beta_1=0.9$, and $\beta_2=0.999$.
We apply cosine learning-rate decay with a linear warm-up of 1{,}000 steps, a peak learning rate of $1\times 10^{-5}$ towards a minimum learning rate of $0.0$.


\section{Evaluation}
\label{sec:evaluation}
\subsection{Experimental Setup}
\label{sec:exp_setup}

\begin{table*}[t]
\centering
\small
\caption{Dataset Statistics Comparison (AIRTAG vs.\ Ours).}
\label{tab:data-stats-airtag-ours}
\setlength{\tabcolsep}{3pt}
\resizebox{\textwidth}{!}{
\begin{tabular}{
l p{3.5cm}
S[table-format=7.0] S[table-format=6.0] S[table-format=2.1]
S[table-format=3.0] c
S[table-format=7.0] S[table-format=6.0] S[table-format=2.1]
S[table-format=3.0] c
}
\toprule
& & \multicolumn{5}{c}{AIRTAG Preprocessed Dataset} & \multicolumn{5}{c}{Our Preprocessed Dataset} \\
\cmidrule(lr){3-7}\cmidrule(lr){8-12}
Split & APT intrusion campaign
& {Total Entries} & {Attack} & {\%} & {\scriptsize Size[MB]} & {char. $\mu \pm \sigma$}
& {Total Entries} & {Attack} & {\%} & {\scriptsize Size[MB]} & {char. $\mu \pm \sigma$} \\
\midrule
M1 & Strategic web compromise
& 251675 & 8180  & 3.3 &  28 & 115.0 $\pm$  66.3
& 274713 & 9007  & 3.3 & 179 & 681.0 $\pm$ 310.1 \\
M2 & Targeted GOV phishing
& 284392 & 34968 & 12.3&  33 & 113.4 $\pm$  64.4
& 311585 & 35171 & 11.3& 208 & 695.9 $\pm$ 331.8 \\
M3 & Malvertising dominate
& 334343 & 35198 & 10.5&  37 & 111.5 $\pm$  57.6
& 358133 & 35123 & 9.8 & 228 & 667.4 $\pm$ 364.4 \\
M4 & Monero miner by Rig
& 258750 & 8260  & 3.2 &  30 & 114.8 $\pm$  62.2
& 282014 & 9319  & 3.3 & 185 & 686.7 $\pm$ 314.5 \\
M5 & Pony campaign
& 701512 & 34256 & 4.9 &  77 & 114.2 $\pm$  61.3
& 767192 & 34313 & 4.5 & 475 & 669.6 $\pm$ 215.5 \\
M6 & Spam campaign
& 354028 & 10004 & 2.8 &  39 & 112.0 $\pm$  59.9
& 380945 & 11485 & 3.0 & 243 & 660.8 $\pm$ 278.5 \\
S1 & Strategic web compromise
&  95065 & 4598  & 4.8 &  11 & 116.0 $\pm$  71.0
& 102213 & 5262  & 5.1 &  65 & 664.8 $\pm$ 281.5 \\
S2 & Malvertising dominate
& 397952 & 15073 & 3.8 &  46 & 117.6 $\pm$  67.2
& 419412 & 17989 & 4.3 & 270 & 672.9 $\pm$ 243.5 \\
S3 & Spam campaign
& 128317 & 5172  & 4.0 &  15 & 118.7 $\pm$  69.2
& 140320 & 5826  & 4.2 &  92 & 679.6 $\pm$ 317.3 \\
S4 & Pony campaign
& 125613 & 18071 & 14.4&  15 & 119.9 $\pm$  70.9
& 135534 & 18137 & 13.4&  89 & 681.1 $\pm$ 315.0 \\
\midrule
Total &
& 2931647 & 173780 & 5.9 &  &
& 3172061 & 181632 & 5.7 &  &  \\
\bottomrule
\end{tabular}
}
\end{table*}

\paragraph{Datasets and preprocessing settings.}
To evaluate the robustness of CAPTAIN with respect to dataset characteristics, we conduct two experiments using datasets produced by different preprocessing pipelines.

\begin{itemize}
    \item \textbf{Experiment 1 (AIRTAG-preprocessed ATLAS).} We use the dataset preprocessed by the AIRTAG/ATLAS pipeline.
    \item \textbf{Experiment 2 (CAPTAIN-preprocessed ATLAS).} We use the dataset produced by our minimal, domain-agnostic preprocessing.
\end{itemize}

A comparison of these two datasets is summarized in Table~\ref{tab:data-stats-airtag-ours}.
Overall, the AIRTAG-preprocessed data has smaller file sizes and shorter log entries: the average entry length is approximately 115 characters (with a standard deviation around 60).
In contrast, our preprocessed data is larger and contains substantially longer entries: the average entry length is approximately 670 characters (with a standard deviation around 300), which indicates that it retains more information per entry.

\paragraph{Experiment 1: Evaluation on AIRTAG-preprocessed data.}
In Experiment 1, we compare AIRTAG and CAPTAIN on the dataset preprocessed by the AIRTAG/ATLAS pipeline.
Since AIRTAG provides a tokenizer specialized for this preprocessed format, we integrate the same tokenizer into the text encoder used by CAPTAIN in this experiment.
Because the text encoder must adapt to the AIRTAG tokenizer in this setting, we also make the text encoder parameters trainable during the adapter training stage (only for this experiment).

We evaluate two maximum input lengths: \textbf{32 tokens}, which is AIRTAG's default setting, and \textbf{64 tokens} as an extended setting to test robustness to a larger token budget.

To match AIRTAG's evaluation protocol, we use the same train/test split configuration for each target dataset.
Specifically, for multi-host campaigns:
(i) when testing on M1 or M2, we train on M3--M6;
(ii) when testing on M3 or M4, we train on M1, M2, M5, and M6;
(iii) when testing on M5 or M6, we train on M1--M4.
For single-host datasets S1--S4, when testing on a dataset (e.g., S1), we train on the remaining ones (e.g., S2--S4).

\paragraph{Experiment 2: Evaluation on CAPTAIN-preprocessed data.}
In Experiment 2, we compare AIRTAG and CAPTAIN on the dataset produced by our preprocessing.
For AIRTAG, we follow the authors' released scripts as closely as possible to build the required vocabulary file, training data, and test data from our dataset.
In this experiment, we adopt the ATLAS-style training configuration for train/test splits.
For example, to test on M1, we train on M2--M6; to test on S1, we train on S2--S4.

\paragraph{Performance metric.}
To compare detection performance, we use the \textbf{area under the curve (AUC) of receiver operating characteristics (ROC)}.
AUC summarizes the trade-off between the TPR and the FPR across all possible decision thresholds $\tau$.

For AIRTAG, the OC-SVM produces a signed score based on the distance of each data point to the learned decision boundary (hyperplane).
We sweep the classification threshold over this score to compute TPR and FPR, and then compute AUC. For CAPTAIN, we use the perplexity-based score produced by the LM for each log entry.
We sweep a threshold $\tau$ over the perplexity values to compute TPR and FPR, and then compute AUC.

In both Experiment~1 and Experiment~2, we observe dataset splits where the AUC falls below 0.5 for both AIRTAG and CAPTAIN. Since an AUC below 0.5 indicates that the scoring function is negatively correlated with the ground-truth labels (i.e., reversing the decision rule would yield better-than-random discrimination), we report the corresponding discriminative ability using $1-\mathrm{AUC}$ for those cases.

\subsection{Experiment 1 Results: Evaluation on
AIRTAG-preprocessed Dataset}
\label{sec:eval_airtag_preprocessed}

We report the results of Experiment~1 in Table~\ref{tab:exp12-comp-results}.
Overall, AIRTAG achieves very strong performance when the maximum input length is 32 tokens, especially on multi-host splits M1--M4.
However, when the maximum input length is increased to 64 tokens, AIRTAG's performance decreases across all splits, resulting in an average AUC drop of $0.16$.

In contrast, CAPTAIN is substantially more stable with respect to the input token budget.
With 32 tokens, CAPTAIN outperforms AIRTAG on M5, M6, S1, S2, S3, S4, and the overall average.
More importantly, when increasing the input length from 32 to 64 tokens, CAPTAIN exhibits only a minor degradation: the average AUC decreases by $0.02$.
These results indicate that CAPTAIN maintains detection performance without relying on a tightly constrained token budget.

\subsection{Experiment 2 Results:
Evaluation on
CAPTAIN-preprocessed Dataset}
\label{sec:eval_ours_preprocessed}

We also report the results of Experiment~2 in Table~\ref{tab:exp12-comp-results}.
On our minimally preprocessed dataset, CAPTAIN outperforms AIRTAG on \textbf{all} splits.
In particular, CAPTAIN maintains a high average AUC of $0.93$, whereas AIRTAG achieves an average AUC of $0.68$.

These results suggest that CAPTAIN does not require heavily engineered preprocessing to achieve strong detection performance.
Even on less curated logs produced by domain-agnostic preprocessing, CAPTAIN remains effective, indicating that it is robust to dataset characteristics and input variability.

We also observe that the AUC on S2 falls below 0.5 for both AIRTAG and CAPTAIN. 
We attribute this to S2 containing a large volume of benign logs that are highly specific to that split, which can skew the data distribution and hinder generalization (Appendix~\ref{apx:unique_logs_on_S2}).

\begin{table*}[t]
  \centering
  \caption{\textbf{AUC comparison on Experiment 1 and Experiment 2.}}
  \label{tab:exp12-comp-results}
  \setlength{\tabcolsep}{4pt}
  \renewcommand{\arraystretch}{1.1}
  \begin{threeparttable}
  \begin{tabular}{crrrrrr}
    \toprule
    & \multicolumn{4}{c}{Experiment 1} & \multicolumn{2}{c}{Experiment 2} \\
    \cmidrule(lr){2-5}\cmidrule(lr){6-7}
    & \multicolumn{2}{c}{32 tokens} & \multicolumn{2}{c}{64 tokens} & \multicolumn{2}{c}{512 tokens} \\
    \cmidrule(lr){2-3}\cmidrule(lr){4-5}\cmidrule(lr){6-7}
    Split & \multicolumn{1}{c}{AIRTAG} & \multicolumn{1}{c}{CAPTAIN} & \multicolumn{1}{c}{AIRTAG} & \multicolumn{1}{c}{CAPTAIN} & \multicolumn{1}{c}{AIRTAG} & CAPTAIN \\
    \midrule

    M1 & \bestbarvalA{0.997} & \barvalC{0.981} & \barvalA{0.959} & \bestbarvalC{0.977} & \barvalA{0.736} & \bestbarvalC{0.951} \\
    M2 & \bestbarvalA{0.999} & \barvalC{0.988} & \barvalA{0.980} & \bestbarvalC{0.985} & \barvalA{0.680} & \bestbarvalC{0.950} \\
    M3 & \bestbarvalA{0.998} & \barvalC{0.994} & \barvalA{0.873} & \bestbarvalC{0.997} & \barvalA{0.874} & \bestbarvalC{0.976} \\
    M4 & \bestbarvalA{0.998} & \barvalC{0.993} & \barvalA{0.775} & \bestbarvalC{0.992} & \barvalA{0.888} & \bestbarvalC{0.966} \\
    M5 & \barvalA{0.943} & \bestbarvalC{0.979} & \barvalA{0.530} & \bestbarvalC{0.975} & *\barvalA{0.541} & \bestbarvalC{0.979} \\
    M6 & \barvalA{0.989} & \bestbarvalC{0.991} & \barvalA{0.936} & \bestbarvalC{0.982} & \barvalA{0.688} & \bestbarvalC{0.971} \\
    S1 & \barvalA{0.972} & \bestbarvalC{0.986} & \barvalA{0.869} & \bestbarvalC{0.994} & \barvalA{0.563} & \bestbarvalC{0.918} \\
    S2 & \barvalA{0.820} & \bestbarvalC{0.979} & *\barvalA{0.592} & \bestbarvalC{0.793} & *\barvalA{0.689} & *\bestbarvalC{0.746} \\
    S3 & \barvalA{0.932} & \bestbarvalC{0.984} & \barvalA{0.616} & \bestbarvalC{0.981} & \barvalA{0.509} & \bestbarvalC{0.888} \\
    S4 & \barvalA{0.934} & \bestbarvalC{0.951} & \barvalA{0.846} & \bestbarvalC{0.935} & \barvalA{0.672} & \bestbarvalC{0.944} \\

    \rowcolor{avebg}
    Ave. & \barvalA{0.958} & \bestbarvalC{0.983} & \barvalA{0.798} & \bestbarvalC{0.961} & \barvalA{0.684} & \bestbarvalC{0.929} \\
    \bottomrule
  \end{tabular}
  \begin{tablenotes}[flushleft]
      \footnotesize
      \item \textit{Note:} AUC values below 0.5 are inverted and marked with *.
    \end{tablenotes}
  \end{threeparttable}
\end{table*}

\subsection{Ablation Study}
\label{sec:ablation}

\begin{figure*}[t]
\centering

\begin{subfigure}[t]{0.49\textwidth}
  \centering
  \includegraphics[width=\linewidth]{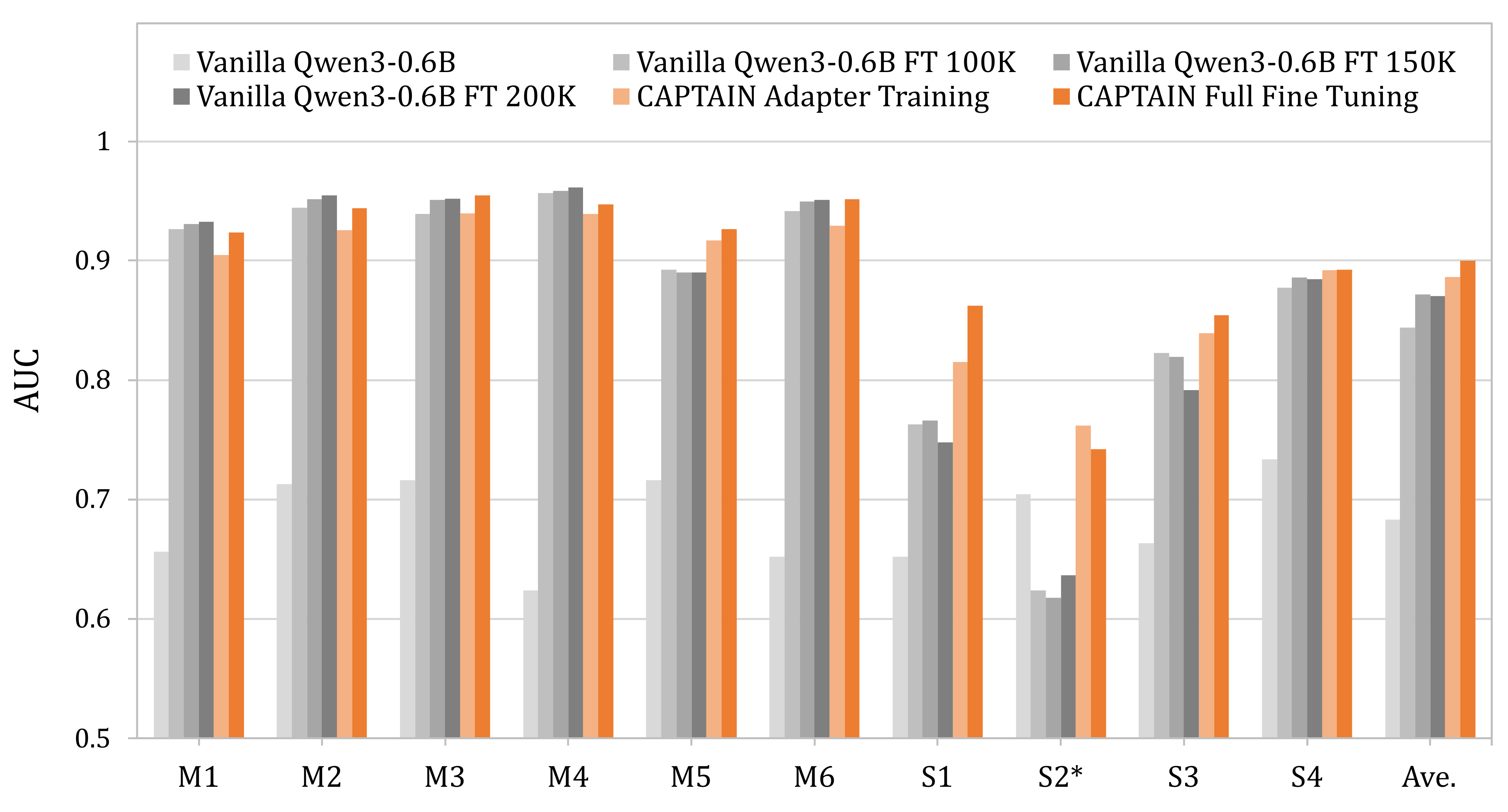}
  \caption{\textbf{Q-Former ablation.}}
  \label{fig:vanilla_diff}
\end{subfigure}\hfill
\begin{subfigure}[t]{0.49\textwidth}
  \centering
  \includegraphics[width=\linewidth]{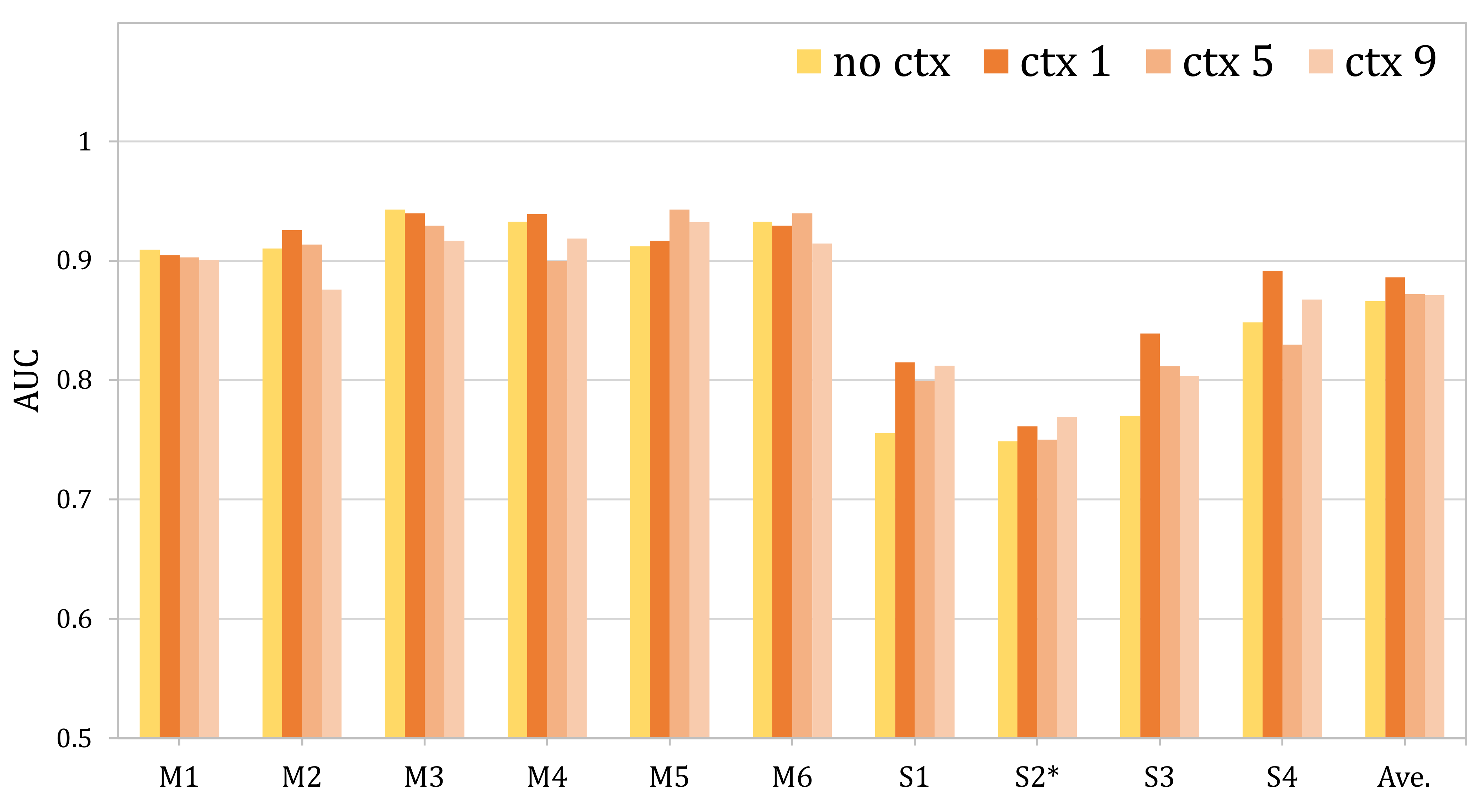}
  \caption{\textbf{Varying context window size.}}
  \label{fig:context_diff}
\end{subfigure}

\vspace{6pt}

\begin{subfigure}[t]{0.49\textwidth}
  \centering
  \includegraphics[width=\linewidth]{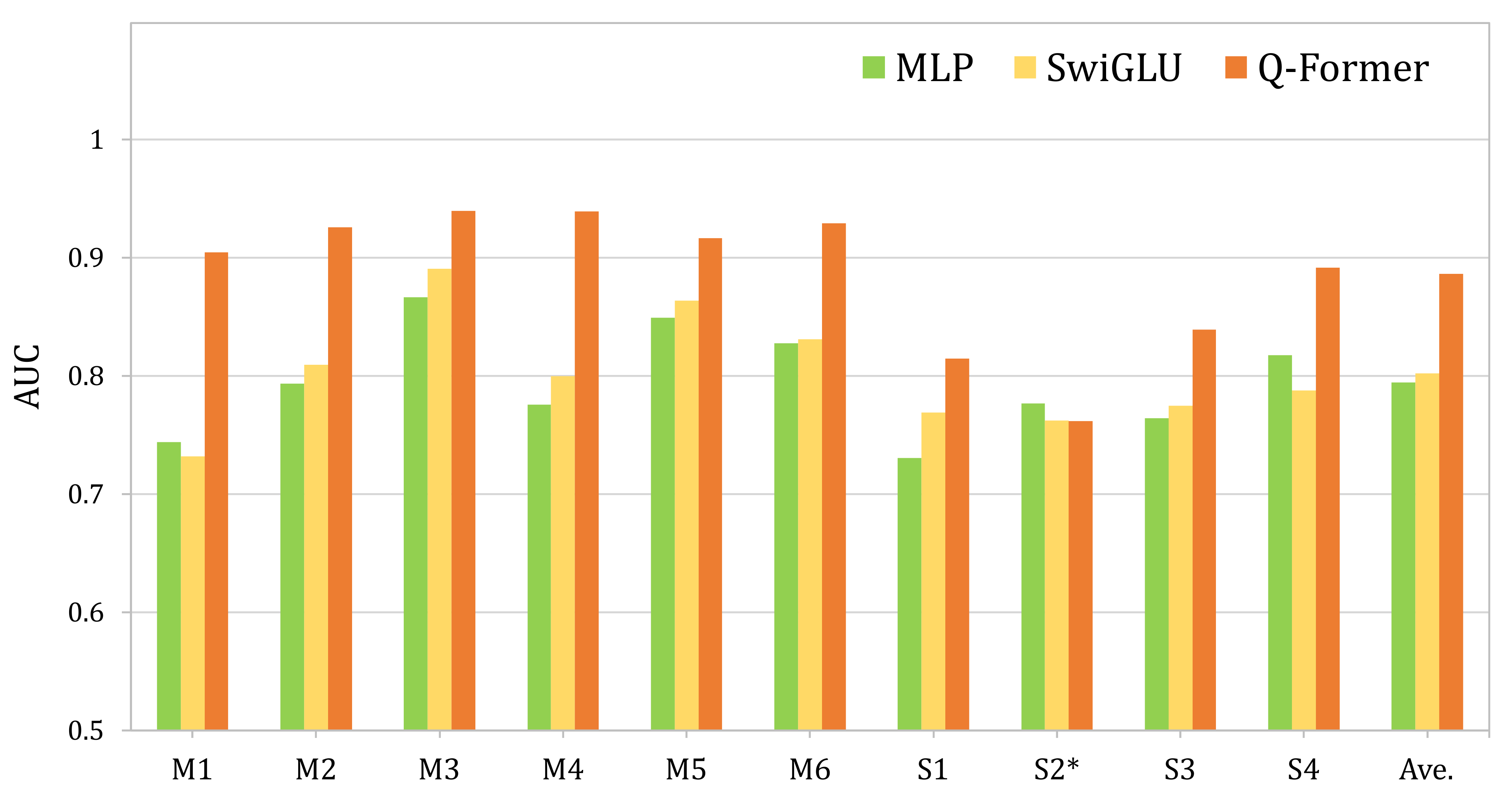}
  \caption{\textbf{Bridge adapter replacement.}}
  \label{fig:adapter_diff}
\end{subfigure}\hfill
\begin{subfigure}[t]{0.49\textwidth}
  \centering
  \includegraphics[width=\linewidth]{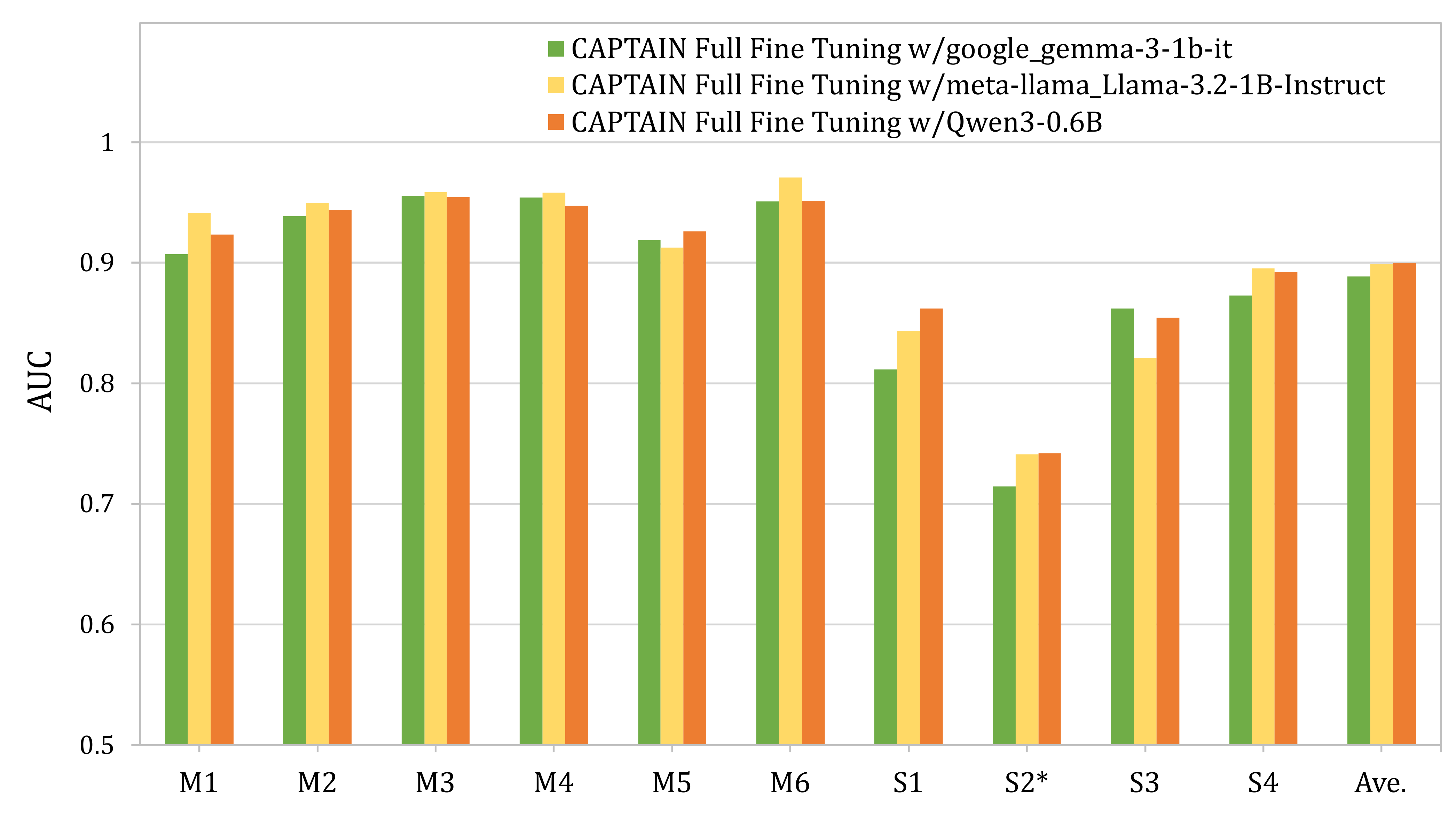}
  \caption{\textbf{Decoder LM replacement.}}
  \label{fig:lm_diff}
\end{subfigure}

\caption{\textbf{Ablation study of Q-former and LM.} We analyze the impact of (a) removing Q-Former (vanilla LM baselines with additional fine-tuning), (b) varying the history context window size, (c) replacing Q-Former with alternative adapters, and (d) swapping the decoder LM backbone.}
\label{fig:ablation}
\end{figure*}

In this section, we conduct an ablation study to isolate the key design choices that contribute to CAPTAIN's performance.
CAPTAIN is built on two major components:
(1) a \textbf{context-bridging mechanism} that incorporates historical log context and connects the encoder space to the decoder LM space via a Q-Former, and
(2) a \textbf{decoder-only LM} that serves as the backbone for PPL-based scoring.

In addition, since post-processing is also an important component that contributes to performance, we include it in our ablation study as well.

Our goal is to quantify how much each component contributes to detection performance and to validate whether the corresponding design choices are justified.

All ablation experiments in this section are conducted on our preprocessed dataset, which contains longer and less curated log entries and therefore better reflects realistic operating conditions.

\subsubsection{Comparison with Vanilla LM}
\label{sec:ablation_no_qformer}

we evaluate the contribution of CAPTAIN's core mechanism: soft prompting via a Q-Former that incorporates historical log context and bridges the encoder space and the decoder LM space.
To this end, we compare CAPTAIN against a vanilla decoder-only LM that does not use Q-Former or contextual soft tokens.
In addition, we fine-tune the vanilla LM using only benign logs in order to separate the effect of simple domain adaptation from the structural benefit introduced by CAPTAIN.

Figure~\ref{fig:ablation}~(a) shows the results.
First, fine-tuning the vanilla LM improves performance, indicating that adapting a decoder-only LM to the log domain is beneficial.
However, even the fine-tuned vanilla LM does not reach the performance of CAPTAIN, even without full fine-tuning of stage 2.

We also observe an additional improvement from after CAPTAIN adapter training (stage 1) to CAPTAIN full fine-tuning (stage 2).
This indicates that it is not sufficient to only calibrate the Q-Former bridge; end-to-end adaptation that jointly fine-tunes the encoder and the decoder LM is also important.

Finally, the effect of increasing the fine-tuning budget for the vanilla LM (100K/150K/200K steps) is not monotonic.
In particular, the model trained for 200K steps performs worse than the one trained for 150K steps.
This highlights that longer training is not always beneficial and can introduce issues such as overfitting, distributional bias, or optimization instability.

\subsubsection{Effect of Historical Context Size}
\label{sec:ablation_context_size}

We next study how the amount of historical context affects detection performance. We vary the context size, defined as the number of past log entries provided to the model.

Figure~\ref{fig:ablation}~(b) summarizes the results on the ATLAS dataset. When the context size is set to 0, 1, 5, and 9, the average AUC follows the order:
1 (AUC: 0.886) $>$ 5 (0.872) $>$ 9 (0.871) $>$ 0 (0.866).
The lowest average AUC is observed when no context is used (context size 0). This confirms that incorporating historical log context is beneficial for detection. However, performance does not improve monotonically as the context size increases.

\subsubsection{Replacing Q-Former with Alternative Adapters}
\label{sec:ablation_adapters}

In this subsection, we evaluate whether Q-Former is an appropriate bridging mechanism by replacing it with alternative adapters. We consider an multi-layer perceptron (MLP) adapter and a SwiGLU-based adapter. We choose these designs because they are widely used as projector/connector modules in recent multimodal models. For example, \cite{liu2023improved} reports that a fully-connected MLP connector between the vision encoder and the LLM can be strong and data-efficient. In addition, recent efficiency-oriented multimodal models explore connector designs that include SwiGLU \cite{noam2020glu, wang2024coglvm}. For instance, ML-Mamba incorporates SwiGLU into the multimodal connector to improve the ingestion of visual sequences \cite{huang2024mlmamba}. Based on these literature, we treat MLP and SwiGLU as strong alternative baselines for CAPTAIN's bridge.

The results in Figure~\ref{fig:ablation}~(c) show that the average AUC follows the order:
Q-Former (0.886) $>$ SwiGLU (0.802) $>$ MLP (0.794).
This ranking suggests that CAPTAIN benefits from more than a simple projection from the encoder space to the LM space.
SwiGLU can improve over an MLP by increasing nonlinearity and expressive capacity via gating, but it still does not compete with Q-Former.

\subsubsection{Effect of Decoder LM Backbone}
\label{sec:ablation_lm_backbone}

We also evaluate whether CAPTAIN depends on a particular decoder-only LM, and how the choice of backbone affects performance. We replace the decoder LM with several lightweight open models that have more parameters than Qwen3-0.6B. Specifically, we evaluate Llama3.2-1B-Instruct \cite{dubey2024llama3herd, meta2024llama3.2} and Gemma3-1B-IT \cite{gemmateam2025gemma3, google_gemma_3_1b_it}.

The results in Figure~\ref{fig:ablation}~(d) show that the average AUC follows the order:
Qwen3-0.6B (0.900) $>$ Llama3.2-1B-IT(0.899) $>$ Gemma3-1B-IT (0.887).
Qwen3-0.6B and Llama3.2-1B-IT achieve almost comparable performance, while Gemma3-1B-IT yields a slightly lower average AUC.

One possible explanation is that these instruction-tuned (IT) models may be optimized for dialogue and instruction-following behavior more strongly than for likelihood-based modeling of domain text such as logs.
In many IT models, the generation distribution is shaped using prompt templates and supervised preferences designed for interactive tasks.
In CAPTAIN, however, our full fine-tuning stage performs domain adaptation on log data without using such instruction-style prompt templates.
Under this training setup, transferring from a highly instruction-tuned distribution to the log-domain distribution may be less effective, which can lead to similar or slightly worse AUC despite having more parameters.
From a deployment perspective, it is also encouraging that Qwen3-0.6B achieves strong performance with a relatively small model size.
This aligns with our target use case: instead of aggregating logs to a central server and relying on large-scale compute, we aim for fast detection on each host with limited resources. The fact that a lightweight decoder LM can deliver competitive accuracy supports the practicality of CAPTAIN.

\subsubsection{Effect of Wiener Filter.}
\label{sec:effect_post-processing}
CAPTAIN includes a post-processing step that smooths the per-entry PPL sequence using a Wiener filter.

\begin{figure}[t]
  \centering
  \includegraphics[width=\linewidth]{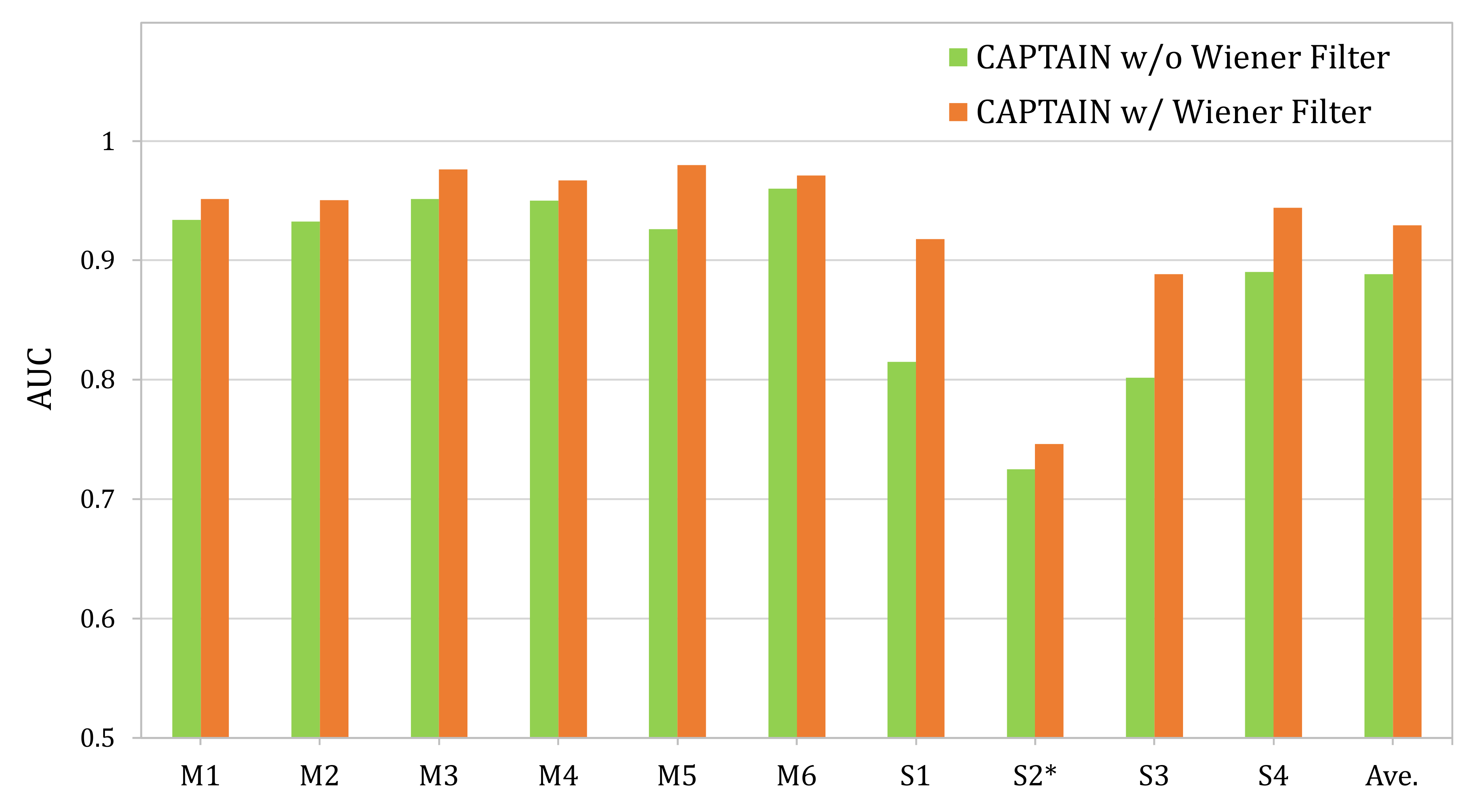}
  \caption{\textbf{Wiener Filter Ablation.}}
  \label{fig:filter_diff}
\end{figure}

Figure~\ref{fig:filter_diff} quantifies the impact of applying the Wiener filter in Experiment~2. Compared to the same model without any smoothing (i.e., the fully fine-tuned CAPTAIN model evaluated on raw PPL scores), Wiener filtering improves performance on all data splits. On average, the AUC increases from 0.889 without filtering to 0.929 with filtering, indicating a consistent benefit from simple temporal smoothing of the PPL time series.

\section{Discussion}
\label{sec:discussion}

\subsection{Robustness}
\label{sec:discussion_robustness}

The results from Experiment~1 and Experiment~2 suggest a clear difference in robustness between AIRTAG and CAPTAIN.
AIRTAG is sensitive to both the input token budget and the preprocessing pipeline, while CAPTAIN remains stable under these changes.

We attribute this difference primarily to how each method represents log entries and performs anomaly detection.
AIRTAG compresses each log entry into a single fixed-size embedding (a 128-dimensional vector corresponding to the \texttt{[CLS]} token) using a compact BERT encoder, and then applies a one-class classifier (OC-SVM).
This design can create a strong representational bottleneck.
The \texttt{[CLS]} token has been widely used as an aggregate representation for downstream tasks \cite{devlin2019bert,choi2021evaluation}, but prior work has reported that \texttt{[CLS]}-based representations can be weak as sentence-level semantic embeddings \cite{reimers2019sentencebert}, and that the behavior of BERT-based classifiers can be sensitive to changes in the input text \cite{jin2019isbert}.
These observations are consistent with a setting where a single-vector summary is not always stable under changes in input length or noise.

In Experiment~1, increasing the token budget (from 32 to 64) adds more trailing tokens that are often less relevant for detection in the curated ATLAS format.
This can increase the variance of \texttt{[CLS]} embeddings even for benign logs, because the encoder must compress more content into a single fixed-size vector.
When benign embeddings become more dispersed, a one-class detector such as OC-SVM may need to cover a larger region of the embedding space, which can reduce separability between benign and malicious samples and degrade the ROC curve.

In Experiment~2, the effect becomes stronger because our minimally preprocessed logs are longer and noisier.
Long entries contain more variability, and local attack indicators can be diluted when mapped into a single fixed-size representation.
Such shifts in the embedding distribution can degrade the separability assumed by OC-SVM and lead to lower AUC (and worse TPR/FPR trade-offs).

In contrast, CAPTAIN uses a decoder-only language model to directly score each log entry by the likelihood of its token sequence (i.e., perplexity).
Perplexity reflects the average cross-entropy between the input sequence and the learned domain distribution.
It does not rely on a particular embedding geometry learned under a specific preprocessing pipeline in the same way as a fixed \texttt{[CLS]}-embedding space.
In addition, CAPTAIN explicitly injects historical log context through context tokens produced by the encoder and the Q-Former bridge.
This encourages the LM to interpret an entry in relation to recent activity, rather than relying only on entry-local keywords.
Finally, CAPTAIN treats the perplexity scores as a time series and applies smoothing.
This reduces sensitivity to noisy individual events and emphasizes sustained deviations, which are more consistent with typical threat activity patterns.

Taken together, these design choices explain why CAPTAIN can maintain performance even when the token budget increases or when preprocessing becomes less curated.

We emphasize that this discussion does not imply that AIRTAG is ineffective; rather, it highlights that a \texttt{[CLS]}-compression plus one-class boundary can be sensitive to input distribution shifts, which becomes visible under more deployment-oriented settings.

\subsection{Effect of Q-Former-based Soft Prompting}
\label{sec:discussion_qformer}

Our ablation results show that the performance gain of CAPTAIN cannot be explained solely by additional training on log data (i.e., domain adaptation). 
Instead, the context bridge enabled by Q-Former---which transfers information from the encoder space to the decoder LM space---plays a central role.

\paragraph{(a) Structural benefit beyond domain adaptation.}
When we remove Q-Former and use a vanilla decoder-only LM, fine-tuning on benign logs improves performance. This confirms that adapting a decoder-only LM to the log domain is beneficial.
However, even the fine-tuned vanilla LM does not match CAPTAIN, especially the adapter trained (stage 1 only) variant.
This gap suggests that the improvement is not explained by domain adaptation alone: projecting contextual information from the encoder space into the decoder LM space through the Q-Former provides a substantial benefit.

The additional gain from CAPTAIN stage 1 (adapter training focused on Q-Former) to stage 2 (end-to-end fine-tuning including the encoder and LM) further supports this interpretation.
It is not enough to make the bridge function in isolation; the encoder and LM must also co-adapt so that the decoder LM can reliably utilize the injected context tokens.
Importantly, this improvement is not replicated by simply training the vanilla LM longer, which strengthens the conclusion that the bridging mechanism and joint optimization are key.

We also observe that increasing the fine-tuning budget of the vanilla LM does not yield monotonic improvements (e.g., 200K steps underperforms 150K steps).
This suggests that, for log data with repetitive structures, learning from individual entries can be prone to overfitting or optimization instability.
In contrast, CAPTAIN conditions on historical context during training, so even similar log entries can appear under different surrounding contexts.
This may encourage the model to rely less on memorizing surface strings and more on context-dependent consistency, which can mitigate overfitting.

\paragraph{(b) Context size and the need for selective information.}
When we vary the context size, using no context (size 0) performs the worst, while a small context (size 1) performs the best.
This indicates that incorporating historical context is useful, but increasing history indiscriminately is not.
Excessive context may act as noise or make training more challenging.

This behavior is consistent with CAPTAIN's design: past logs are compressed into a fixed number of learned queries (context tokens) before being consumed by the LM.
As the number of past entries $N$ increases, compression becomes more severe and important cues may be averaged out or diluted.
Therefore, while context provides a clear advantage over no-context scoring, too much history can be counterproductive, highlighting the importance of selective extraction.

\paragraph{(c) Adapter replacement: projection alone is not sufficient.}
Replacing Q-Former with MLP or SwiGLU adapters yields the ranking Q-Former $>$ SwiGLU $>$ MLP in average AUC.
This suggests that CAPTAIN benefits not only from mapping features into the LM space, but also from selectively extracting and shaping useful fragments from past logs via cross-attention.
An MLP applies a single global mapping, and SwiGLU can increase expressiveness through gating, which explains its improvement over MLP.
However, neither matches Q-Former, suggesting that increased nonlinearity alone cannot replace the selectivity provided by attention-based bridging. 
In other words, higher expressiveness by itself appears insufficient; effective detection likely requires selectively extracting relevant signals via attention.

Overall, these results indicate that while the choice of context size matters, Q-Former is a crucial design element of CAPTAIN,
enabling effective context-aware perplexity scoring by providing a selective and LM-compatible representation of historical log context.

\subsection{Limitations and Future Work}
\label{sec:limitations}

We discuss several limitations of this work and outline directions for future research.

\paragraph{Dataset dependence.}
Our evaluation is conducted primarily on ATLAS.
Although ATLAS is constructed from multiple log sources and includes a range of injected attack behaviors, it remains a benchmark dataset and cannot cover the full diversity of real operational environments.
Therefore, our results do not guarantee the same level of performance under different logging configurations, software stacks, or organizational practices.
Evaluating CAPTAIN on additional public datasets and on logs collected from real deployments is an important next step to assess generalization from multiple perspectives.

\paragraph{The ``anomaly = attack'' gap under unlabeled training.}
CAPTAIN is trained using only benign logs in a one-class/unsupervised manner.
As a result, legitimate but previously unseen changes in system behavior (e.g., operational updates or workload shifts) can also yield high perplexity and be flagged as suspicious.
In fact, on S2 we observe an AUC below 0.5, which indicates that benign entries can receive higher perplexity than malicious ones in that split.
In such a case, reversing the decision direction (treating low perplexity as anomalous) would yield a higher detection rate.
We attribute this behavior to the presence of a large volume of benign log patterns that appear only in S2.
Notably, we observe a similar phenomenon for AIRTAG in our evaluation on our preprocessed dataset (Section~4, Experiment~2), suggesting that this challenge is not unique to CAPTAIN but reflects a broader issue in one-class detection for logs.
Future work may address benign distribution shifts through adaptive thresholding, continual learning, or lightweight human-in-the-loop feedback to separate operational changes from true attacks.

\paragraph{Uncertainty in context design and the optimal window size.}
As shown in our ablation study, historical context improves detection performance, but excessive context can introduce noise and degrade performance.
The optimal context size may vary depending on the environment, log granularity, and the time scale of attacks.
Developing adaptive context selection strategies is therefore an important direction, such as selecting salient events, introducing hierarchical memory representations, or dynamically adjusting the number of context queries.

\paragraph{Broader applicability and multimodality.}
Since CAPTAIN detects out-of-distribution behavior through perplexity, it may be applicable not only to APT detection but also to other anomaly detection tasks.
In addition, CAPTAIN uses Q-Former to bridge representations from a text encoder into a decoder LM space.
Originally, Q-Former was proposed to project visual embeddings from a vision encoder into an LM-compatible space, and recent work has highlighted the importance of multimodal signals for anomaly detection.
A natural extension is to incorporate additional modalities beyond text.
For example, time-series telemetry such as network traffic volume could be transformed into visual summaries (e.g., plots) and encoded with a vision encoder, enabling the model to detect anomalies from complementary perspectives.
Exploring such multimodal extensions, as well as applications to other anomaly detection settings, is a promising future direction.

\section{Related Work}
\label{sec:related}

\paragraph{Robust Log Anomaly Detection with Machine Learning}
A wide range of machine learning techniques, including deep learning, have been studied for anomaly detection in software logs from systems and applications \cite{landauer2023deep,yadav2020asurvey,chen2022experiencere,siwach2022anomaly,xu2025deeplearning}.
Early approaches often relied on engineered representations and assumptions about log formats.
Representative examples include mining-based detectors that exploit invariants across log events \cite{lou2010mining}, PCA-based methods that detect outliers by measuring distances from the principal subspace of normal behavior \cite{xu2009detecting}, and feature-based classifiers that perform binary classification with manually constructed features \cite{he2018towards,liang2007failure}.
A common limitation of these approaches is that they can be sensitive to changes in log formats and templates.

To improve robustness to log variation, learning-based methods that capture the semantics and ordering of log messages have been proposed.
LogRobust, for example, learns semantic representations and sequence patterns to detect anomalies under changing log templates \cite{zhang2019robust}.
However, LogRobust is a supervised method, and thus it inherits practical challenges such as the scarcity of high-quality labeled data and limited portability across environments.

\paragraph{Deep Learning for Attack Detection from Audit Logs}
In security settings, deep learning has also been applied to attack detection from audit logs, often by constructing provenance graphs (causal graphs) from low-level events and learning over graph-derived sequences.
Representative systems include ATLAS \cite{alsaheel2021atlas}, CLUE \cite{cui2025clue}, and ConLBS \cite{li2023conlbs}.
These approaches can achieve high detection accuracy, but they typically require generating and manipulating provenance graphs, which can introduce substantial computational overhead and delay time-to-detection.
In addition, abstracting raw logs into graph nodes and edges can reduce granularity, and prior work reports that such abstraction may lead to unstable or inaccurate results in some settings \cite{ding2023airtag}.

In contrast, CAPTAIN directly analyzes log entries without requiring provenance graph construction, enabling faster detection while preserving the original granularity of log evidence.
This design is aligned with operational settings where low latency and lightweight deployment are important.

\paragraph{LLM-based Detection and Incident Analysis}
Recently, applying large language models (LLMs) to security tasks such as attack detection and incident analysis has attracted growing attention \cite{hassanin2024comprehensive, xu2025large, hasanov2024application}.
Several methods leverage LLMs' language understanding and reasoning capabilities to improve detection or to support higher-level investigation workflows \cite{sun2025fromalert,wang2026logppo,meng2025knowhow}.
CLOUSEAU \cite{aldaihan2025clouseau}, in particular, proposes a hierarchical multi-agent system that uses LLM reasoning to automate cyber attack investigation and reconstruct attack narratives without relying on task-specific training or hand-crafted rules.
It reduces noise by extracting only events relevant to a user-provided Point-of-Interest (POI) and then tracing causal relations in a step-by-step manner.

CLOUSEAU targets a complementary use case to ours: it assumes a POI as an initial hint, whereas CAPTAIN aims to identify suspicious log entries without such prior guidance.
Accordingly, the two approaches can coexist in a practical pipeline, e.g., CAPTAIN can first surface candidate malicious logs, and CLOUSEAU can then expand the analysis to reconstruct the attack story around these findings.

Many LLM-based systems also require substantial computational resources, which can be a barrier for host-level deployment.
CAPTAIN, by contrast, is designed to operate with relatively lightweight open-source encoder and decoder transformers, enabling detection on each host under limited computational resources.

\section{Conclusion}
\label{sec:conclusion}

This paper proposed CAPTAIN, a context-augmented, perplexity-based approach for detecting threat activities from logs.
We first revisited a strong baseline, AIRTAG, and re-evaluated it under settings that remove factors that may unintentionally affect the observed performance.
Motivated by practical limitations of pipelines that rely on heavily curated, entry-level log representations, we then introduced several mechanisms to tackle them.

CAPTAIN scores log entries using perplexity from a decoder-only language model and injects historical context via an encoder and a Q-Former bridge.
By additionally treating perplexity scores as a time series and applying smoothing, CAPTAIN achieves robustness to changes in token budget and preprocessing, and it remains effective on long and noisy logs produced by minimal preprocessing.

In experiments on the ATLAS dataset, CAPTAIN competes well with AIRTAG under the original curated setting while exhibiting substantially more stable performance under increased token budgets and altered preprocessing.
CAPTAIN also consistently achieved high performance on less curated logs that are longer and noisier.
Our ablation study showed that Q-Former-based soft prompting contributes beyond simple domain adaptation, and that effective detection is possible even with lightweight decoder LMs.
Taken together, these results suggest that CAPTAIN is a practical and robust approach for log-based threat detection.

\cleardoublepage
\appendix

\section*{Ethical Considerations}

This work studies automated threat-activity detection from audit and network logs and proposes CAPTAIN, a context-augmented perplexity-based detector. We structure this discussion as a stakeholder-based analysis, distinguishing impacts arising from (i) the research process (data handling and experimentation) and (ii) the publication of results (deployment and broader use).

\paragraph{Stakeholders.}
The primary stakeholders include:
(1) operators and defenders who deploy log-based detection in enterprises;
(2) users and employees whose activity may be recorded in system logs;
(3) the security research community that builds on, evaluates, or reuses our artifacts; and
(4) adversaries who may attempt to evade detection or misuse the method.

\paragraph{Research process: data handling and experimentation.}
Our experiments use the publicly available ATLAS dataset and do not involve interactions with human subjects, active attacks, or collection of new operational telemetry.
Nevertheless, logs can contain sensitive information (e.g., hostnames, user identifiers, filenames, command lines). To reduce privacy risks, we (i) rely on a dataset that is already curated for research distribution, (ii) avoid attempting to enrich the dataset with external sources, and (iii) ensure that our preprocessing does not introduce additional identifying information beyond what is already present in the released corpus.

\paragraph{Publication and deployment impacts: benefits.}
CAPTAIN can benefit defenders by reducing the manual effort required to triage large volumes of logs and by providing a detection approach that remains effective under less curated preprocessing. Improving the reliability of threat detection can also reduce the time-to-response and limit the impact of real attacks.

\paragraph{Publication and deployment impacts: potential harms.}
We consider several potential negative outcomes:
(1) \emph{False positives and operational disruption:} as a one-class method trained on benign data, CAPTAIN can flag benign distribution shifts as anomalies, which may lead to unnecessary investigations or disruptions if used without safeguards.
(2) \emph{Misuse by adversaries:} public descriptions of detection pipelines can help attackers understand what signals may trigger alerts and motivate evasion strategies.
(3) \emph{Over-reliance:} organizations may over-trust automated scores and under-invest in complementary controls and human review.

\paragraph{Mitigations.}
We take and recommend the following mitigations to reduce these risks:
(1) \emph{Data minimization and access control:} deployments should limit log collection and retention to what is necessary for security operations, and restrict access to logs and model outputs to authorized personnel.
(2) \emph{Human-in-the-loop triage:} CAPTAIN should be used as a prioritization signal; high-scoring events should be reviewed in context, especially after operational changes.
(3) \emph{Calibration under distribution shift:} deployments should re-calibrate thresholds when workloads, logging policies, or software versions change.
(4) \emph{Responsible artifact release:} when releasing code and scripts, we avoid including any operational credentials or environment-specific identifiers. we also document intended use.

\paragraph{Decision to conduct and publish.}
We believe the expected benefits to defenders and to the research community outweigh the foreseeable harms, provided the above mitigations are followed. Our study is based on an open research dataset and focuses on robustness of threat detection. We emphasize appropriate governance, access control, and human oversight in any real deployment.

\cleardoublepage
\bibliographystyle{plain}
\bibliography{refs}

\begin{thebibliography}{10}

\bibitem{abrar2025onthereproducibility}
Talha Abrar, Ahmad Shamail, Mohammad~Jaffer Iqbal, Amaan Ahmed, Muhammad Abdullah, Muhammad Shayan, Fareed Zaffar, Thomas Pasquier, David Eyers, and Ashish Gehani.
\newblock On the reproducibility of provenance-based intrusion detection that uses deep learning.
\newblock In {\em Proceedings of the 3rd ACM Conference on Reproducibility and Replicability}, ACM REP '25, page 14–28, New York, NY, USA, 2025. Association for Computing Machinery.

\bibitem{aldaihan2025clouseau}
Abdullah Aldaihan, Fahad Alotaibi, and Sergio Maffeis.
\newblock Clouseau: A hierarchical multi-agent approach for autonomous attack investigation.
\newblock In {\em Annual Computer Security Applications Conference}, 2025.

\bibitem{alsaheel2021atlas}
Abdulellah Alsaheel, Yuhong Nan, Shiqing Ma, Le~Yu, Gregory Walkup, Z.~Berkay Celik, Xiangyu Zhang, and Dongyan Xu.
\newblock {ATLAS}: A sequence-based learning approach for attack investigation.
\newblock In {\em 30th USENIX Security Symposium (USENIX Security 21)}, pages 3005--3022. USENIX Association, August 2021.
\newblock \url{https://www.usenix.org/conference/usenixsecurity21/presentation/alsaheel}.

\bibitem{brown2020language}
Tom Brown, Benjamin Mann, Nick Ryder, Melanie Subbiah, Jared~D Kaplan, Prafulla Dhariwal, Arvind Neelakantan, Pranav Shyam, Girish Sastry, Amanda Askell, Sandhini Agarwal, Ariel Herbert-Voss, Gretchen Krueger, Tom Henighan, Rewon Child, Aditya Ramesh, Daniel Ziegler, Jeffrey Wu, Clemens Winter, Chris Hesse, Mark Chen, Eric Sigler, Mateusz Litwin, Scott Gray, Benjamin Chess, Jack Clark, Christopher Berner, Sam McCandlish, Alec Radford, Ilya Sutskever, and Dario Amodei.
\newblock Language models are few-shot learners.
\newblock In H.~Larochelle, M.~Ranzato, R.~Hadsell, M.F. Balcan, and H.~Lin, editors, {\em Advances in Neural Information Processing Systems}, volume~33, pages 1877--1901. Curran Associates, Inc., 2020.

\bibitem{chen2022experiencere}
Zhuangbin Chen, Jinyang Liu, Wenwei Gu, Yuxin Su, and Michael~R. Lyu.
\newblock Experience report: Deep learning-based system log analysis for anomaly detection, 2022.

\bibitem{cheng2023kairos}
Zijun Cheng, Qiujian Lv, Jinyuan Liang, Yan Wang, Degang Sun, Thomas Pasquier, and Xueyuan Han.
\newblock Kairos: Practical intrusion detection and investigation using whole-system provenance.
\newblock In {\em 2024 IEEE Symposium on Security and Privacy (SP)}, pages 3533--3551, 2024.

\bibitem{choi2021evaluation}
Hyunjin Choi, Judong Kim, Seongho Joe, and Youngjune Gwon.
\newblock { Evaluation of BERT and ALBERT Sentence Embedding Performance on Downstream NLP Tasks }.
\newblock In {\em 2020 25th International Conference on Pattern Recognition (ICPR)}, pages 5482--5487, Los Alamitos, CA, USA, January 2021. IEEE Computer Society.

\bibitem{cui2025clue}
Wenzhuo Cui, Maihao Guo, Jingjing Feng, Shuyi Zhang, Zheng Liu, and Yu~Wen.
\newblock Clue: A high-performance, efficient, and robust apt detection framework via fine-tuning pretrained transformer and contrastive learning.
\newblock In {\em 2025 International Conference on Intelligent Computing}, 2025.

\bibitem{google_gemma_3_1b_it}
Google DeepMind.
\newblock gemma-3-1b-it (gemma 3 1b instruction-tuned) model card.
\newblock \url{https://huggingface.co/google/gemma-3-1b-it}.
\newblock Accessed: 2026-01-29.

\bibitem{devlin2019bert}
Jacob Devlin, Ming{-}Wei Chang, Kenton Lee, and Kristina Toutanova.
\newblock {BERT:} pre-training of deep bidirectional transformers for language understanding.
\newblock {\em CoRR}, abs/1810.04805, 2018.

\bibitem{ding2023airtag}
Hailun Ding, Juan Zhai, Yuhong Nan, and Shiqing Ma.
\newblock {AIRTAG}: Towards automated attack investigation by unsupervised learning with log texts.
\newblock In {\em 32nd USENIX Security Symposium (USENIX Security 23)}, pages 373--390, Anaheim, CA, August 2023. USENIX Association.
\newblock \url{https://www.usenix.org/conference/usenixsecurity23/presentation/ding-hailun-airtag}.

\bibitem{du2017deeplog}
Min Du, Feifei Li, Guineng Zheng, and Vivek Srikumar.
\newblock Deeplog: Anomaly detection and diagnosis from system logs through deep learning.
\newblock In {\em Proceedings of the 2017 ACM SIGSAC Conference on Computer and Communications Security}, CCS '17, page 1285–1298, New York, NY, USA, 2017. Association for Computing Machinery.

\bibitem{dubey2024llama3herd}
Abhimanyu Dubey et~al.
\newblock The llama 3 herd of models, 2024.

\bibitem{pengcheng2022backpropagation}
Pengcheng Fang, Peng Gao, Changlin Liu, Erman Ayday, Kangkook Jee, Ting Wang, Yanfang~(Fanny) Ye, Zhuotao Liu, and Xusheng Xiao.
\newblock {Back-Propagating} system dependency impact for attack investigation.
\newblock In {\em 31st USENIX Security Symposium (USENIX Security 22)}, pages 2461--2478, Boston, MA, August 2022. USENIX Association.
\newblock \url{https://www.usenix.org/conference/usenixsecurity22/presentation/fang}.

\bibitem{guo2021logbert}
Haixuan Guo, Shuhan Yuan, and Xintao Wu.
\newblock Logbert: Log anomaly detection via bert.
\newblock In {\em 2021 International Joint Conference on Neural Networks (IJCNN)}, pages 1--8, 2021.

\bibitem{han2023loggpt}
Xiao Han, Shuhan Yuan, and Mohamed Trabelsi.
\newblock Loggpt: Log anomaly detection via gpt.
\newblock In {\em 2023 IEEE International Conference on Big Data (BigData)}, pages 1117--1122, 2023.

\bibitem{hasanov2024application}
Ismayil Hasanov, Seppo Virtanen, Antti Hakkala, and Jouni Isoaho.
\newblock Application of large language models in cybersecurity: A systematic literature review.
\newblock {\em IEEE Access}, 12:176751--176778, 2024.

\bibitem{hassanin2024comprehensive}
Mohammed Hassanin and Nour Moustafa.
\newblock A comprehensive overview of large language models (llms) for cyber defences: Opportunities and directions, 2024.

\bibitem{he2018towards}
Pinjia He, Jieming Zhu, Shilin He, Jian Li, and Michael~R. Lyu.
\newblock Towards automated log parsing for large-scale log data analysis.
\newblock {\em IEEE Transactions on Dependable and Secure Computing}, 15(6):931--944, 2018.

\bibitem{huang2024mlmamba}
Wenjun Huang, Jiakai Pan, Jiahao Tang, Yanyu Ding, Yifei Xing, Yuhe Wang, Zhengzhuo Wang, and Jianguo Hu.
\newblock Ml-mamba: Efficient multi-modal large language model utilizing mamba-2, 2024.

\bibitem{jia2024magic}
Zian Jia, Yun Xiong, Yuhong Nan, Yao Zhang, Jinjing Zhao, and Mi~Wen.
\newblock {MAGIC}: Detecting advanced persistent threats via masked graph representation learning.
\newblock In {\em 33rd USENIX Security Symposium (USENIX Security 24)}, pages 5197--5214, Philadelphia, PA, August 2024. USENIX Association.
\newblock \url{https://www.usenix.org/conference/usenixsecurity24/presentation/jia-zian}.

\bibitem{jin2019isbert}
Di~Jin, Zhijing Jin, Joey~Tianyi Zhou, and Peter Szolovits.
\newblock Is bert really robust? a strong baseline for natural language attack on text classification and entailment.
\newblock In {\em AAAI Conference on Artificial Intelligence}, 2019.

\bibitem{landauer2023deep}
Max Landauer, Sebastian Onder, Florian Skopik, and Markus Wurzenberger.
\newblock Deep learning for anomaly detection in log data: A survey.
\newblock {\em Machine Learning with Applications}, 12:100470, June 2023.

\bibitem{lee2023lanobert}
Yukyung Lee, Jina Kim, and Pilsung Kang.
\newblock Lanobert: System log anomaly detection based on bert masked language model.
\newblock {\em Appl. Soft Comput.}, 146(C), October 2023.

\bibitem{lewis2020bart}
Mike Lewis, Yinhan Liu, Naman Goyal, Marjan Ghazvininejad, Abdelrahman Mohamed, Omer Levy, Veselin Stoyanov, and Luke Zettlemoyer.
\newblock {BART}: Denoising sequence-to-sequence pre-training for natural language generation, translation, and comprehension.
\newblock In Dan Jurafsky, Joyce Chai, Natalie Schluter, and Joel Tetreault, editors, {\em Proceedings of the 58th Annual Meeting of the Association for Computational Linguistics}, pages 7871--7880, Online, July 2020. Association for Computational Linguistics.

\bibitem{li2023conlbs}
Jiawei Li, Ru~Zhang, and Jianyi Liu.
\newblock Conlbs: An attack investigation approach using contrastive learning with behavior sequence.
\newblock {\em Sensors}, 23(24), 2023.

\bibitem{li2023blip2}
Junnan Li, Dongxu Li, Silvio Savarese, and Steven Hoi.
\newblock Blip-2: Bootstrapping language-image pre-training with frozen image encoders and large language models, 2023.

\bibitem{liang2007failure}
Yinglung Liang, Yanyong Zhang, Hui Xiong, and Ramendra Sahoo.
\newblock Failure prediction in ibm bluegene/l event logs.
\newblock In {\em Seventh IEEE International Conference on Data Mining (ICDM 2007)}, pages 583--588, 2007.

\bibitem{lin2016logclustering}
Qingwei Lin, Hongyu Zhang, Jian-Guang Lou, Yu~Zhang, and Xuewei Chen.
\newblock Log clustering based problem identification for online service systems.
\newblock In {\em 2016 IEEE/ACM 38th International Conference on Software Engineering Companion (ICSE-C)}, pages 102--111, 2016.

\bibitem{liu2023improved}
Haotian Liu, Chunyuan Li, Yuheng Li, and Yong~Jae Lee.
\newblock Improved baselines with visual instruction tuning.
\newblock In {\em 2024 IEEE/CVF Conference on Computer Vision and Pattern Recognition (CVPR)}, pages 26286--26296, 2024.

\bibitem{lou2010mining}
Jian-Guang Lou, Qiang Fu, Shengqi Yang, Ye~Xu, and Jiang Li.
\newblock Mining invariants from console logs for system problem detection.
\newblock In {\em Proceedings of the 2010 USENIX Conference on USENIX Annual Technical Conference}, USENIXATC'10, page~24, USA, 2010. USENIX Association.

\bibitem{lv2024trec}
Mingqi Lv, Hongzhe Gao, Xuebo Qiu, Tieming Chen, Tiantian Zhu, Jinyin Chen, and Shouling Ji.
\newblock Trec: Apt tactic / technique recognition via few-shot provenance subgraph learning.
\newblock In {\em Proceedings of the 2024 on ACM SIGSAC Conference on Computer and Communications Security}, CCS '24, page 139–152, New York, NY, USA, 2024. Association for Computing Machinery.

\bibitem{meng2025knowhow}
Yuhan Meng, Shaofei Li, Jiaping Gui, Peng Jiang, and Ding Li.
\newblock Knowhow: Automatically applying high-level cti knowledge for interpretable and accurate provenance analysis, 2025.

\bibitem{meta2024llama3.2}
Meta.
\newblock Llama 3.2 1b instruct, 2024.
\newblock Model card and weights release. Release date: Sept 25, 2024. \url{https://huggingface.co/meta-llama/Llama-3.2-1B-Instruct}.

\bibitem{ouyang2022training}
Long Ouyang, Jeffrey Wu, Xu~Jiang, Diogo Almeida, Carroll Wainwright, Pamela Mishkin, Chong Zhang, Sandhini Agarwal, Katarina Slama, Alex Ray, et~al.
\newblock Training language models to follow instructions with human feedback.
\newblock {\em Advances in neural information processing systems}, 35:27730--27744, 2022.

\bibitem{raffel2020exploring}
Colin Raffel, Noam Shazeer, Adam Roberts, Katherine Lee, Sharan Narang, Michael Matena, Yanqi Zhou, Wei Li, and Peter~J. Liu.
\newblock Exploring the limits of transfer learning with a unified text-to-text transformer.
\newblock {\em J. Mach. Learn. Res.}, 21(1), January 2020.

\bibitem{reimers2019sentencebert}
Nils Reimers and Iryna Gurevych.
\newblock Sentence-bert: Sentence embeddings using siamese bert-networks.
\newblock {\em ArXiv}, abs/1908.10084, 2019.

\bibitem{sanh2019distilbert}
Victor Sanh, Lysandre Debut, Julien Chaumond, and Thomas Wolf.
\newblock Distilbert, a distilled version of bert: smaller, faster, cheaper and lighter.
\newblock {\em ArXiv}, abs/1910.01108, 2019.

\bibitem{noam2020glu}
Noam Shazeer.
\newblock {GLU} variants improve transformer.
\newblock {\em CoRR}, abs/2002.05202, 2020.

\bibitem{siwach2022anomaly}
Meena Siwach and Suman Mann.
\newblock Anomaly detection for web log data analysis: A review.
\newblock {\em Journal of Algebraic Statistics}, 13(1), 2022.

\bibitem{sun2025fromalert}
Danyu Sun, Jinghuai Zhang, Jiacen Xu, Yu~Zheng, Yuan Tian, and Zhou Li.
\newblock From alerts to intelligence: A novel llm-aided framework for host-based intrusion detection, 2025.

\bibitem{gemmateam2025gemma3}
Gemma Team et~al.
\newblock Gemma 3 technical report, 2025.

\bibitem{vaswani2017attention}
Ashish Vaswani, Noam Shazeer, Niki Parmar, Jakob Uszkoreit, Llion Jones, Aidan~N Gomez, \L~ukasz Kaiser, and Illia Polosukhin.
\newblock Attention is all you need.
\newblock In I.~Guyon, U.~Von Luxburg, S.~Bengio, H.~Wallach, R.~Fergus, S.~Vishwanathan, and R.~Garnett, editors, {\em Advances in Neural Information Processing Systems}, volume~30. Curran Associates, Inc., 2017.

\bibitem{wang2024coglvm}
Weihan Wang, Qingsong Lv, Wenmeng Yu, Wenyi Hong, Ji~Qi, Yan Wang, Junhui Ji, Zhuoyi Yang, Lei Zhao, Xixuan Song, Jiazheng Xu, Keqin Chen, Bin Xu, Juanzi Li, Yuxiao Dong, Ming Ding, and Jie Tang.
\newblock Cogvlm: Visual expert for pretrained language models.
\newblock In A.~Globerson, L.~Mackey, D.~Belgrave, A.~Fan, U.~Paquet, J.~Tomczak, and C.~Zhang, editors, {\em Advances in Neural Information Processing Systems}, volume~37, pages 121475--121499. Curran Associates, Inc., 2024.

\bibitem{wang2026logppo}
Zhihao Wang, Jiachen Dong, and Chuanchuan Yang.
\newblock Logppo: A log-based anomaly detector aided with proximal policy optimization algorithms.
\newblock {\em Smart Cities}, 9(1), 2026.

\bibitem{wei2022cot}
Jason Wei, Xuezhi Wang, Dale Schuurmans, Maarten Bosma, Brian Ichter, Fei Xia, Ed~H. Chi, Quoc~V. Le, and Denny Zhou.
\newblock Chain-of-thought prompting elicits reasoning in large language models.
\newblock In {\em Proceedings of the 36th International Conference on Neural Information Processing Systems}, NIPS '22, Red Hook, NY, USA, 2022. Curran Associates Inc.

\bibitem{xu2025large}
Hanxiang Xu, Shenao Wang, Ningke Li, Kailong Wang, Yanjie Zhao, Kai Chen, Ting Yu, Yang Liu, and Haoyu Wang.
\newblock Large language models for cyber security: A systematic literature review.
\newblock {\em ACM Trans. Softw. Eng. Methodol.}, September 2025.
\newblock Just Accepted.

\bibitem{xu2009detecting}
Wei Xu, Ling Huang, Armando Fox, David Patterson, and Michael~I. Jordan.
\newblock Detecting large-scale system problems by mining console logs.
\newblock In {\em Proceedings of the ACM SIGOPS 22nd Symposium on Operating Systems Principles}, SOSP '09, page 117–132, New York, NY, USA, 2009. Association for Computing Machinery.

\bibitem{xu2025deeplearning}
Zhiwei Xu, Yujuan Wu, Shiheng Wang, Jiabao Gao, Tian Qiu, Ziqi Wang, Hai Wan, and Xibin Zhao.
\newblock Deep learning-based intrusion detection systems: A survey, 2025.

\bibitem{yadav2020asurvey}
Rakesh~Bahadur Yadav, P~Santosh Kumar, and Sunita~Vikrant Dhavale.
\newblock A survey on log anomaly detection using deep learning.
\newblock In {\em 2020 8th International Conference on Reliability, Infocom Technologies and Optimization (Trends and Future Directions) (ICRITO)}, pages 1215--1220, 2020.

\bibitem{yang2025qwen3}
An~Yang, Anfeng Li, Baosong Yang, Beichen Zhang, Binyuan Hui, Bo~Zheng, Bowen Yu, Chang Gao, Chengen Huang, Chenxu Lv, Chujie Zheng, Dayiheng Liu, Fan Zhou, Fei Huang, Feng Hu, Hao Ge, Haoran Wei, Huan Lin, Jialong Tang, Jian Yang, Jianhong Tu, Jianwei Zhang, Jianxin Yang, Jiaxi Yang, Jing Zhou, Jingren Zhou, Junyang Lin, Kai Dang, Keqin Bao, Kexin Yang, Le~Yu, Lianghao Deng, Mei Li, Mingfeng Xue, Mingze Li, Pei Zhang, Peng Wang, Qin Zhu, Rui Men, Ruize Gao, Shixuan Liu, Shuang Luo, Tianhao Li, Tianyi Tang, Wenbiao Yin, Xingzhang Ren, Xinyu Wang, Xinyu Zhang, Xuancheng Ren, Yang Fan, Yang Su, Yichang Zhang, Yinger Zhang, Yu~Wan, Yuqiong Liu, Zekun Wang, Zeyu Cui, Zhenru Zhang, Zhipeng Zhou, and Zihan Qiu.
\newblock Qwen3 technical report, 2025.

\bibitem{zhang2019robust}
Xu~Zhang, Yong Xu, Qingwei Lin, Bo~Qiao, Hongyu Zhang, Yingnong Dang, Chunyu Xie, Xinsheng Yang, Qian Cheng, Ze~Li, Junjie Chen, Xiaoting He, Randolph Yao, Jian-Guang Lou, Murali Chintalapati, Furao Shen, and Dongmei Zhang.
\newblock Robust log-based anomaly detection on unstable log data.
\newblock In {\em Proceedings of the 2019 27th ACM Joint Meeting on European Software Engineering Conference and Symposium on the Foundations of Software Engineering}, ESEC/FSE 2019, page 807–817, New York, NY, USA, 2019. Association for Computing Machinery.

\end{thebibliography}

\FloatBarrier
\clearpage
\appendix

\section{Additional Experiments of AIRTAG Post-Filtering}
\label{apx:post_filter}

AIRTAG applies a post-filter to remove evidently benign entries from the set of suspicious logs predicted by OC-SVM.
The intended effect is to reduce the false positive rate (FPR) while maintaining a high true positive rate (TPR).
Concretely, the post-filter marks a log entry as malicious if it contains at least one word that satisfies both of the following conditions:
\begin{itemize}
    \item it appears infrequently in the document (i.e., it falls outside the top 30\% by frequency, under the default setting), and
    \item it appears at least eight times within the document.
\end{itemize}
If a log entry contains even one such word, it is classified as malicious; otherwise, it is classified as benign.
The list of frequent terms used by this filter is extracted from ATLAS-preprocessed data (or data with a similar format).

\begin{figure*}[t]
\centering

\begin{minipage}[t]{0.48\linewidth}
  \centering
  \includegraphics[width=\linewidth]{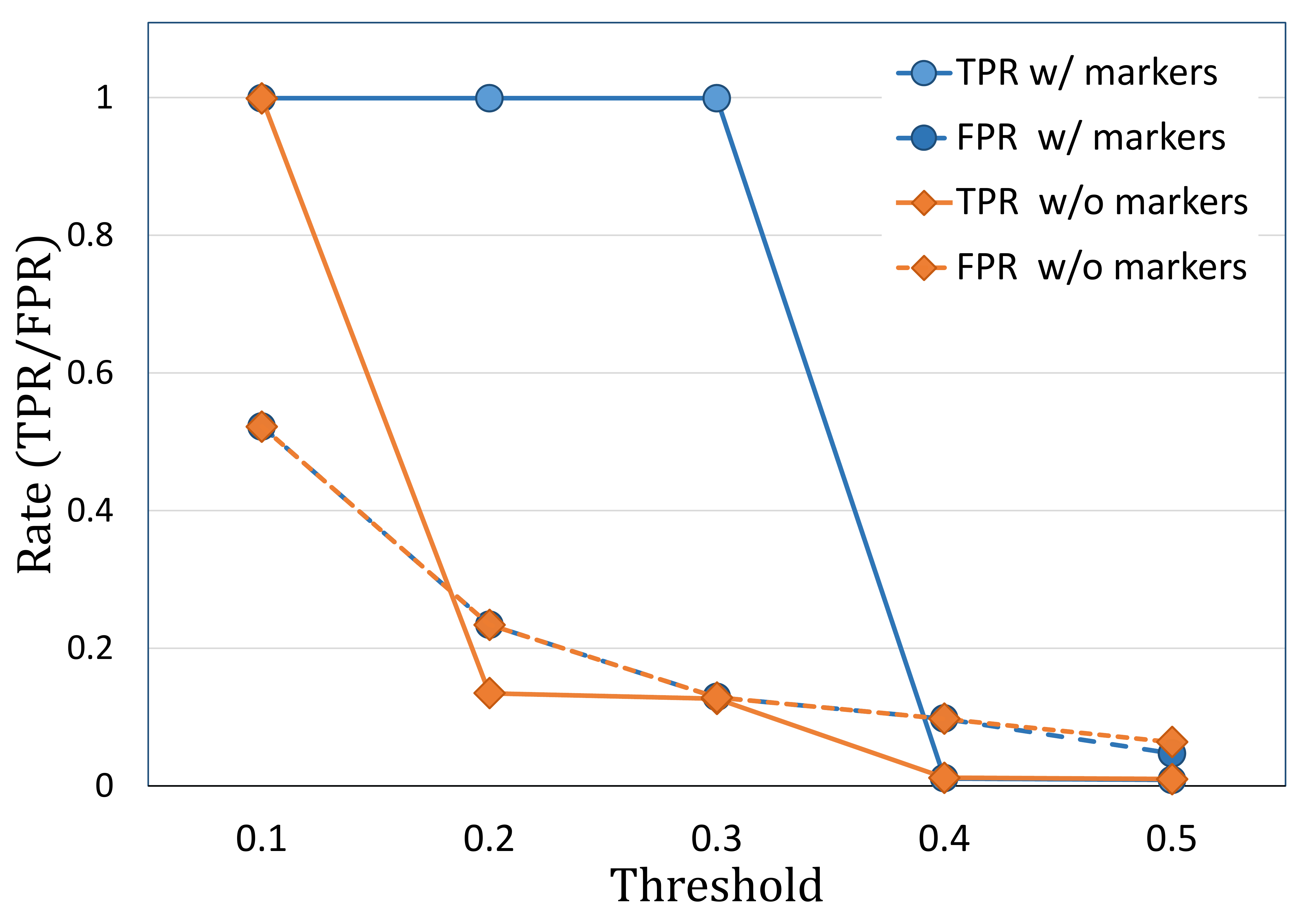}\par
  \vspace{2pt}
  {\small (a) \textbf{S1}}
\end{minipage}\hfill
\begin{minipage}[t]{0.48\linewidth}
  \centering
  \includegraphics[width=\linewidth]{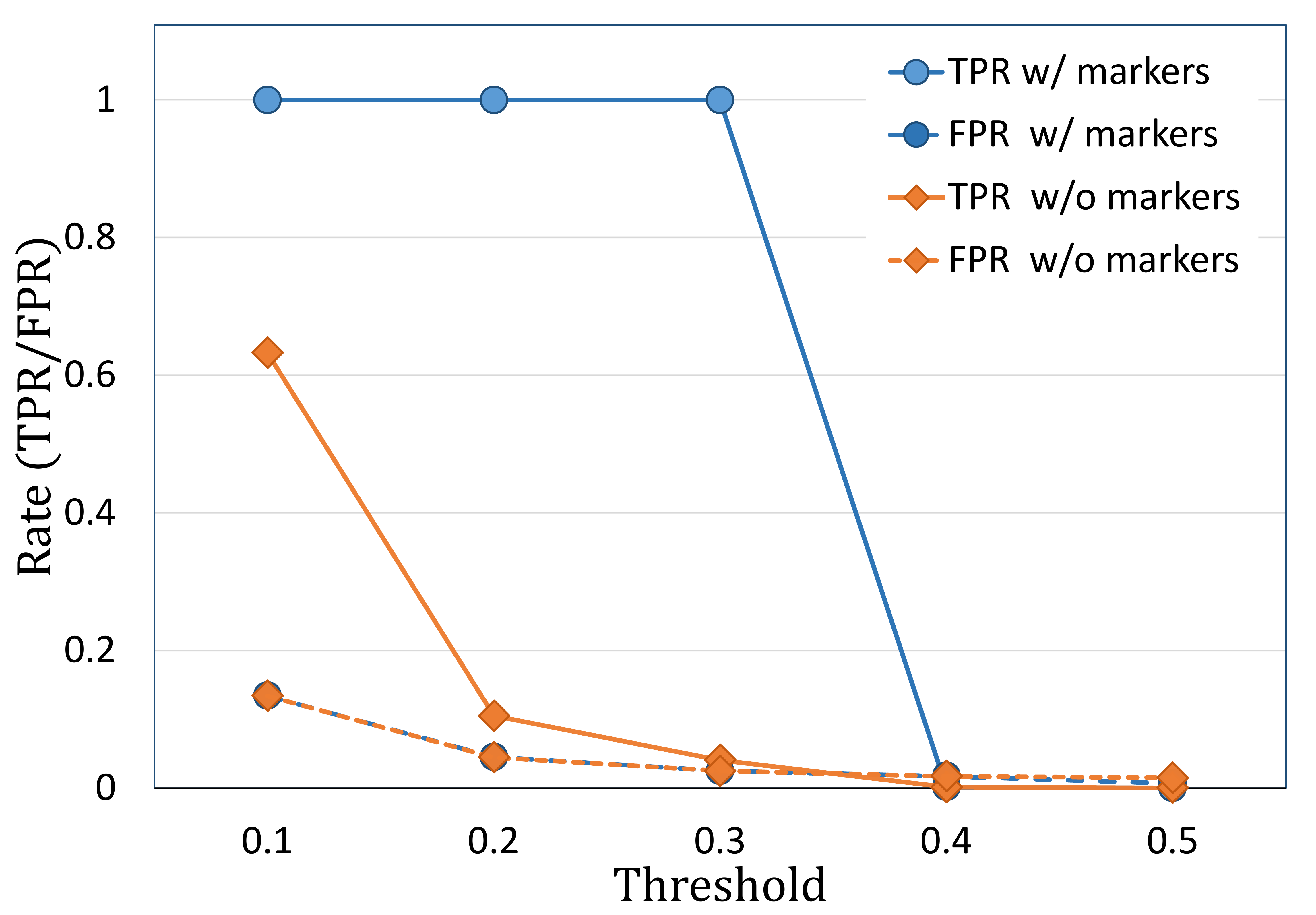}\par
  \vspace{2pt}
  {\small (b) \textbf{S2}}
\end{minipage}

\vspace{6pt}

\begin{minipage}[t]{0.48\linewidth}
  \centering
  \includegraphics[width=\linewidth]{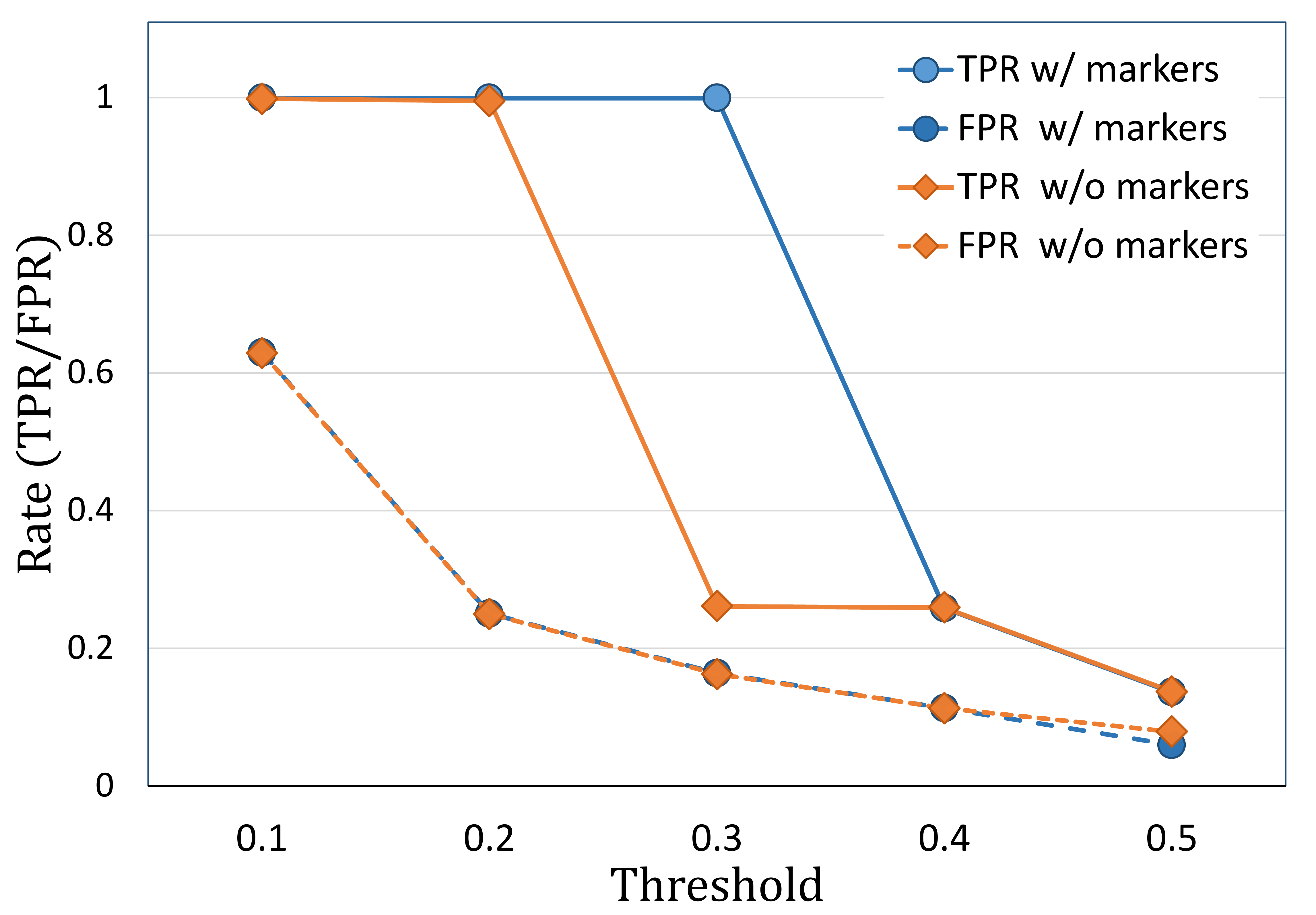}\par
  \vspace{2pt}
  {\small (c) \textbf{S3}}
\end{minipage}\hfill
\begin{minipage}[t]{0.48\linewidth}
  \centering
  \includegraphics[width=\linewidth]{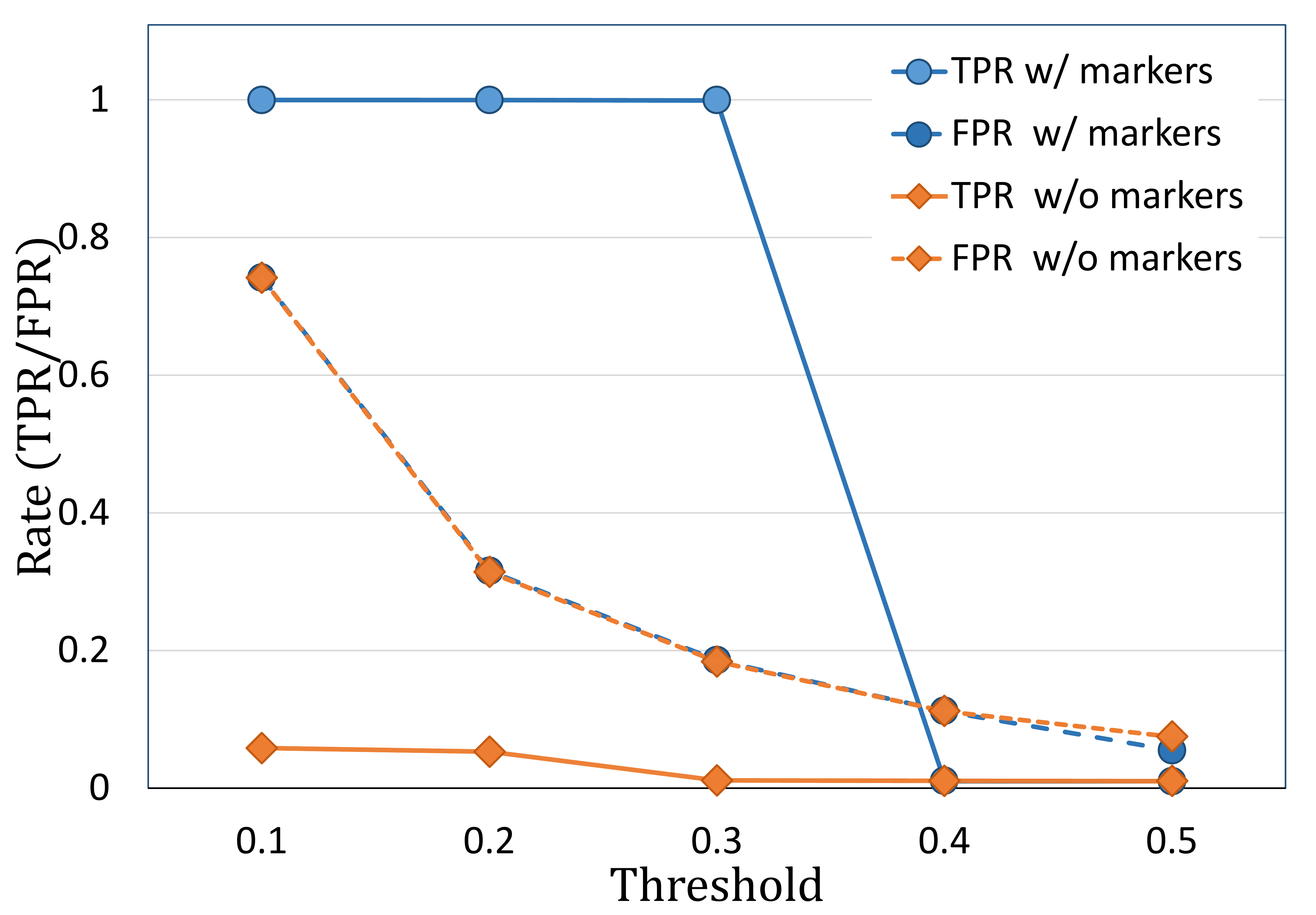}\par
  \vspace{2pt}
  {\small (d) \textbf{S4}}
\end{minipage}

\caption{Threshold–TPR/FPR relationship.}
\label{fig:th-rate-relationship}
\end{figure*}

The AIRTAG paper reports that setting the threshold to $0.3$ consistently reduces FPR while keeping TPR high across datasets.
However, in our investigation, we found that the files used for frequent-term counting provided with the AIRTAG artifacts still contain the ground-truth label markers ``+/-''.
To assess the impact of this artifact, we conducted additional experiments where we removed the ``+/-'' markers and re-measured the post-filter performance on the S1--S4 splits.
Figure~\ref{fig:th-rate-relationship} shows that, after removing the markers, the TPR falls below $0.3$ at threshold $0.3$ for all splits (orange diamond, solid line).
Moreover, within the threshold range of $0.1$--$0.5$, we did not find a common threshold that simultaneously maintains a high TPR while effectively reducing only the FPR.

We also revisit the threshold sensitivity described in the original AIRTAG paper: while a threshold of $0.3$ reduces FPR with little impact on TPR, increasing the threshold beyond $0.4$ leads to a drop in TPR.
This trend is consistent with our measurements using the released artifacts (blue circle, solid line).
Our analysis suggests the following mechanism.
In ATLAS-preprocessed data, the final field of each entry encodes both the log type identifier and the benign/malicious label information.
For example, strings such as ``-LA-'' and ``-LA+'' indicate the source type (e.g., Windows Audit logs) together with the ground-truth label (benign vs.\ malicious).
We confirmed that when the threshold is set to $0.3$, ``-LA-'' is selected as a frequent term, whereas when the threshold is increased to $0.4$, ``-LA+'' becomes selected as a frequent term.
Since the filter treats log entries containing selected frequent terms as benign, selecting ``-LA+'' causes malicious Windows Audit entries to be filtered out as benign, which directly reduces the TPR.

Taken together, these results indicate that the post-filter does not reliably provide the intended benefit under properly sanitized conditions.
Therefore, we exclude this post-filtering step in the remainder of our evaluation.

\section{Temporal bias of attack logs in the ATLAS dataset}
\label{apx:temporal_bias}

We model the binary attack label sequence $\{x_t\}_{t=1}^T$ as a first-order Markov process:
\begin{equation}
P(x_{t+1}\mid x_{1:t}) = P(x_{t+1}\mid x_t), \quad x_t \in \{\mathrm{F}, \mathrm{T}\}.
\label{eq:markov1}
\end{equation}
Here, we treat \texttt{False} as benign logs and \texttt{True} as malicious logs.
Under this assumption, the temporal dependency of attack labels is fully characterized by the
$2\times 2$ transition probabilities $P(\mathrm{F}\mid\mathrm{F})$, $P(\mathrm{T}\mid\mathrm{F})$,
$P(\mathrm{F}\mid\mathrm{T})$, and $P(\mathrm{T}\mid\mathrm{T})$.

To summarize the degree of temporal persistence (i.e., clustering) and switching behavior, we report
the following aggregated quantities:
\begin{equation}
\begin{aligned}
\text{Self-transition} \;&=\; \frac{P(\mathrm{F}\mid \mathrm{F}) + P(\mathrm{T}\mid \mathrm{T})}{2},\\
\text{Switching} \;&=\; \frac{P(\mathrm{T}\mid \mathrm{F}) + P(\mathrm{F}\mid \mathrm{T})}{2}.
\end{aligned}
\label{eq:self_switch}
\end{equation}

A higher self-transition indicates that labels tend to persist across time steps, whereas a higher
switching rate indicates frequent alternations between $\mathrm{F}$ and $\mathrm{T}$.

Table~\ref{tab:markov} shows consistently high self-transition probabilities across all splits
($0.972$--$0.996$) with correspondingly low switching rates ($0.004$--$0.028$). This indicates that,
under a first-order Markov view, the label dynamics are strongly ``sticky'': once the process is in
either $\mathrm{F}$ or $\mathrm{T}$, it is likely to remain in the same state at the next time step.
Such behavior implies substantial temporal clustering in attack logs, rather than rapid alternation.

\begin{table}[t]
  \centering
  \caption{First-order Markov transition probabilities and derived statistics.}
  \label{tab:markov}
  \setlength{\tabcolsep}{6pt}
  \renewcommand{\arraystretch}{1.15}

  \begin{tabular}{lcc}
    \toprule
    Split &
    Self-transition &
    Switching \\
    \midrule
    M1 & 0.980 & 0.020 \\
    M2 & 0.994 & 0.006 \\
    M3 & 0.995 & 0.005 \\
    M4 & 0.978 & 0.022 \\
    M5 & 0.993 & 0.007 \\
    M6 & 0.975 & 0.025 \\
    S1 & 0.977 & 0.023 \\
    S2 & 0.996 & 0.004 \\
    S3 & 0.972 & 0.028 \\
    S4 & 0.994 & 0.006 \\
    \bottomrule
  \end{tabular}
\end{table}

\section{Unique log entries on S2 dataset}
\label{apx:unique_logs_on_S2}
During evaluation, we observed a notable performance drop on the S2 split.
To understand this behavior, we inspected the logs in S2 and identified at least three types of benign log entries that are heavily skewed toward S2 and consistently yield high perplexity.

Table~\ref{tab:s2_biased_logs} lists these three log types (identified by characteristic strings) and their occurrence counts across S1--S4.
In our preprocessed dataset, there are 419{,}412 log entries in total on S2 (Table~\ref{tab:data-stats-airtag-ours}), and we find that more than half of the entries are dominated by logs that appear almost exclusively in S2.
These entries often contain seemingly random strings within the message body (see the example in Figure~\ref{fig:high-ppl-log-entry-on-s2}), which plausibly increases perplexity.

Because these log patterns are largely absent from the other splits (S1, S3, and S4), a model trained without S2 cannot learn that they are benign.
As a result, when evaluated on S2, the model assigns high perplexity to these unseen-but-legitimate entries, leading to degraded detection performance.

\begin{table}[t]
\centering
\small
\caption{Counts of logs disproportionately present in S2.}
\label{tab:s2_biased_logs}
\setlength{\tabcolsep}{6pt}
\begin{tabular}{lrrrr}
\toprule
\textbf{Keyword} & \textbf{S1} & \textbf{S2} & \textbf{S3} & \textbf{S4} \\
\midrule
Audit Policy Change        & 72  & 146629 & 77 & 73 \\
Sensitive Privilege Use    & 157 & 65512  & 1  & 1  \\
31bf3856ad364e35\_         & 0   & 83225  & 3  & 3  \\
\bottomrule
\end{tabular}
\end{table}

\begin{figure}[t]
\centering
\caption{\textbf{Sample Log Entry with High Perplexity}}
\label{fig:high-ppl-log-entry-on-s2}
\begin{lstlisting}
Audit Success   9/15/2018 8:38:54 AM    Microsoft-Windows-Security-Auditing     4674    Sensitive Privilege Use "An operation was attempted on a privileged object.

Subject:
         Security ID:            WIN-D65GVM5K5FO\aalsahee
         Account Name:           aalsahee
         Account Domain:         WIN-D65GVM5K5FO
         Logon ID:               0x19224

Object:^M
         Object Server:  Security
         Object Type:    File
         Object Name:    C:\$Recycle.Bin\S-1-5-21-450080267-1945256726-3465656282-1000\$II85YSE.2924
         Object Handle:  0x31c

Process Information:
         Process ID:     0xbf4
         Process Name:   C:\Windows\System32\dllhost.exe

Requested Operation:
         Desired Access: READ_CONTROL
                                 ACCESS_SYS_SEC
                                 ReadAttributes
                                 
         Privileges:             SeSecurityPrivilege
\end{lstlisting}
\end{figure}

\end{document}